\documentclass[10pt,twocolumn,letterpaper]{article}

\usepackage[rgb]{xcolor}
\usepackage{3dv}
\usepackage{times}
\usepackage{epsfig}
\usepackage{graphicx}
\usepackage{amsmath}
\usepackage{amssymb}

\usepackage{booktabs}
\usepackage{microtype}
\usepackage{xargs}
\usepackage{colortbl}
\definecolor{mygray}{cmyk}{0, 0, 0, 0.55}
\definecolor{kit-blue}{RGB}{70, 100, 170}
\definecolor{kit-blue100}{RGB}{70, 100, 170}
\definecolor{kit-blue50}{rgb}{0.6373, 0.6961, 0.8333}
\definecolor{kit-green}{RGB}{0, 150, 130}
\definecolor{kit-red}{RGB}{162, 34, 35}
\definecolor{kit-orange}{RGB}{223, 155, 27}
\definecolor{kit-yellow}{RGB}{252, 229, 0}
\definecolor{kit-purple}{RGB}{163, 16, 124}
\definecolor{kit-cyan}{RGB}{35, 161, 224}

\definecolor{MYDETAIL}{named}{kit-blue}

\usepackage{mathtools}
\usepackage{siunitx}
\usepackage{bm}
\usepackage{multirow}
\usepackage{tikz} 
\usepackage{pgfplots}
\pgfplotsset{compat=newest}
\usepackage{tikzscale}
\usetikzlibrary{
	angles,
	arrows.meta,
	backgrounds,
	calc,
	decorations.pathreplacing,
	decorations.markings,
	external,
	fit,
	intersections,
	plotmarks,
	positioning,
	shapes,
	quotes,
}
\newcommand{\tikzsize}{\small}
\newcommand{\myarrowhead}{{Latex[length=1mm,width=1mm]}} 

\DeclareMathOperator{\diag}{diag}  

\newcommand{\IfLowercaseGreek}[3]{
	\normalexpandarg
	\IfSubStr{\alpha\beta\gamma\delta\epsilon\varepsilon\zeta\eta\theta\vartheta\iota\kappa\lambda\mu\nu\xi\pi\varpi\rho\varrho\sigma\varsigma\tau\upsilon\phi\varphi\chi\psi\omega}{#1}
	{#2}
	{#3}
}
\newcommand{\IfUppercaseGreek}[3]{
	\normalexpandarg
	\IfSubStr{\Gamma\Delta\Theta\Lambda\Xi\Pi\Sigma\Upsilon\Phi\Psi\Omega\II}{#1}
	{#2}
	{#3}
}
\newcommand{\IfGreek}[3]{
	\IfLowercaseGreek{#1}{#2}{\IfUppercaseGreek{#1}{#2}{#3}}
}
\newcommand{\myvec}[1]{
	\IfLowercaseGreek{#1}{\boldsymbol{#1}}{\boldsymbol{\mathbf{#1}}}
}
\newcommand{\mytensor}[1]{
	\IfLowercaseGreek{#1}{\boldsymbol{\mathcal{#1}}}{\boldsymbol{\mathbf{\mathcal{#1}}}}
}

\usepackage{array}
\newcolumntype{C}[1]{>{\centering\let\newline\\\arraybackslash\hspace{0pt}}m{#1}}
\newcolumntype{L}[1]{>{\raggedright\let\newline\\\arraybackslash\hspace{0pt}}m{#1}}

\usepackage{makecell}

\usepackage[english]{babel}
\usepackage{egabbrev}

\usepackage[pagebackref=true,breaklinks=true,letterpaper=true,colorlinks,bookmarks=false]{hyperref}
\usepackage{amsfonts}

\threedvfinalcopy 

\def\threedvPaperID{314} 

\begin{document}

\title{Spectral Reconstruction and Disparity from\\ Spatio-Spectrally Coded Light Fields via Multi-Task Deep Learning}

\author{\begin{minipage}{5cm}\centering Maximilian Schambach\\[-1mm]{\tt\small schambach@kit.edu}\end{minipage}
	\qquad
	\begin{minipage}{5cm}\centering Jiayang Shi\\[-1mm]{\tt\small jiayang.shi@student.kit.edu}\end{minipage}
	\qquad
	\begin{minipage}{5cm}\centering Michael Heizmann\\[-1mm]{\tt\small michael.heizmann@kit.edu}\end{minipage}\\[4mm]
	Karlsruhe Institute of Technology, Germany
}

\maketitle

\begin{abstract}
	We present a novel method to reconstruct a spectral central view and its aligned disparity map from spatio-spectrally coded light fields.
	Since we do not reconstruct an intermediate full light field from the coded measurement, we refer to this as principal reconstruction.
	The coded light fields correspond to those captured by a light field camera in the unfocused design with a spectrally coded microlens array.
	In this application, the spectrally coded light field camera can be interpreted as a single-shot spectral depth camera.

	We investigate several multi-task deep learning methods and propose a new auxiliary loss-based training strategy to enhance the reconstruction performance.
	The results are evaluated using a synthetic as well as a new real-world spectral light field dataset that we captured using a custom-built camera.
	The results are compared to state-of-the art compressed sensing reconstruction and disparity estimation.

	We achieve a high reconstruction quality for both synthetic and real-world coded light fields.
	The disparity estimation quality is on par with or even outperforms state-of-the-art disparity estimation from uncoded RGB light fields.
\end{abstract}
%
%
%
\vspace{-5mm}
\section{Introduction}\label{sec:introduction}
Compared to traditional images, light fields contain much more information of a captured scene.
Beside the conventional spatial information, light fields additionally capture the scene's angular information which enables applications such as refocusing~\cite{Ng2005}, segmentation~\cite{Wanner2013b}, and saliency detection~\cite{Li2014}, as well as depth or disparity estimation~\cite{Wanner2012,Wanner2013a}.
Moreover, one can extract reflectance properties such as the specular and diffuse components of a scene~\cite{Alperovich2018,Sulc2016}.
However, light fields also show a strong redundancy~\cite{Levoy1996}.
This is particularly true for hand-held plenoptic cameras due to the inherently small baseline.
The redundancy becomes even more severe in the case of multi- or hyperspectral
light fields in which the community has shown an increased interest.
In some instances, spectral light fields outperform conventional RGB light fields, \eg in depth estimation in specular regions~\cite{Zhou2020} or profilometry~\cite{Farber2018:spectral-lf}.
Futhermore, spectral light fields offer new possibilities, combining methods from conventional light field imaging (\eg disparity estimation) with those from spectral imaging (\eg material classification).

\begin{figure}
	\centering
	\input{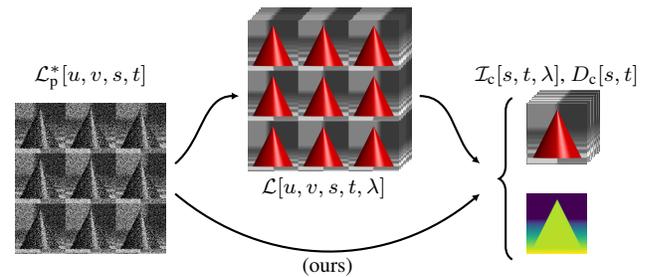}\vspace{-4mm}
	\caption{Schematic comparison of conventional (top) and principal reconstruction (bottom) from a spatio-spectrally coded light field.}
	\label{fig:comparison-traditional-principal}\vspace{-4mm}
\end{figure}

Traditionally, light fields are captured using either a camera array (or gantry), or using a camera with a microlens array (MLA) for which there are two design choices: the unfocused design~\cite{Ng2005} and the focused design~\cite{Lumsdaine2009}.
In the case of spectral light field imaging, various approaches have been proposed, such as multi-camera setups~\cite{Zhu2018,Zhou2020}, single camera snapshot imagers~\cite{Horstmeyer2009,Ye2015,Xiong2017}, as well as a camera using catadioptric mirrors~\cite{Xue2017}.
Despite many benefits accomplished with these cameras, one might ask if it is actually necessary to capture a full light field in order to extract the desired information.
Instead, one can capture a coded light field.
While compressive light field cameras~\cite{Ashok2010,Marwah2013,Pegard2016} as well as compressive spectral light field cameras~\cite{Xiong2017,Marquez2020:Compressive-spectral-lf-image-reconstruction} have been studied in several instances, the focus of these presentations lay on reconstruction using compressed sensing (CS): given a compressed measurement, one can reconstruct the original signal using the $l_0$-norm as a reconstruction prior.
In the case of coded light fields, it seems superfluous to reconstruct the high dimensional light field from the low dimensional compressed measurement, only to extract again low dimensional (yet complex) information from it.
Furthermore, the performance of common light field tasks may suffer due to the errors introduced by the light field reconstruction.
Finally, CS reconstruction is computationally and memory demanding.
Therefore, we propose to reconstruct the desired information from the coded light field directly.
We refer to this as \emph{principal reconstruction}.
Here, we reconstruct the spectral central view and its disparity map but other reconstruction targets are also possible.
We choose these targets as they represent a large amount of the information contained in the full light field.
In fact, for non-occluding Lambertian scenes, the central view and its disparity are equivalent to the full light field.
We perform the reconstruction of coded light fields as taken by a light field camera in the unfocused design with a spectrally coded MLA.

Since neural networks have become the state-of-the-art in estimating intrinsics or disparities from light fields~\cite{Alperovich2018,Shin2018:EPINET} but also in reconstruction from compressed measurements~\cite{Gupta2017,Kabkab2018,Kulkarni2016, Chandramouli2020:Generative-model-lf-reconstructions}, we perform the proposed reconstruction using a supervised deep learning approach.
The estimation of multiple targets from the coded measurements is inherently a multi-task problem.
We investigate several training strategies to enhance the naive multi-task approach which is based on a simple weighted sum of losses.
Furthermore, we propose a new strategy to regularize the single task losses to optimize secondary target metrics via gradient similarity which can be combined with the existing multi-task-adapted training strategies.
In particular, our contributions are as follows:
\begin{itemize}
	\setlength\itemsep{0pt}
	\item We estimate a spectral central view and its disparity map from spatio-spectrally coded light fields as taken by a light field camera with a spectrally coded MLA.
	\item We propose a new training strategy leveraging auxiliary losses which we show enhances existing multi-task approaches in our application.
	\item Using a custom-built camera, we captured a reference dataset containing 12 spectral light fields of three highly textured scenes.
	\item The reconstruction is evaluated using synthetic and real-world data and compared to state-of-the-art CS-based reconstruction with subsequent disparity estimation.
\end{itemize}
The dataset and code are made publicly available~\cite{Schambach:data:dataset,Schambach:git:code}.
\section{Coded Light Fields}
\label{sec:coded-light-fields}
In the plane-plane parametrization, the spectral light field is denoted by $\mathcal{L}(u, v, s, t, \lambda)$, where $(u, v)$ corresponds to the angular component of the light field, $(s, t)$ to the spatial domain, and $\lambda$ denotes the spectral channel index.
For MLA-based light field cameras, $(u, v)$ and $(s, t)$ are typically chosen to parameterize the main lens plane and the MLA plane, respectively.
In the discrete case, one may equivalently use a tensor-based notation and identify
\begin{equation}
	\mathcal{L}[u, v, s, t, \lambda] = \myvec{\mathcal{L}}_{uvst\lambda}\,,\quad \myvec{\mathcal{L}} \in \mathbb{R}^{U\times V\times S\times T\times \Lambda}\,,
\end{equation}
where $(U, V)$ corresponds to the angular, $(S, T)$ to the spatial, and $\Lambda$ to the spectral resolution of the light field.

\textbf{Related works:}
There are several possibilities to code light fields: in the angular, the spatial or the combined spatio-angular domain~\cite{Ashok2010}.
In CS applications, the light field is typically coded in the spatio-angular domain using an attenuation mask~\cite{Marwah2013,Veeraraghavan2007, Vadathya2019:Unified-learning-based-framework}.
However, these methods usually do not explicitly account for the color or spectral domain of the light field and perform the reconstruction channel-wise, employing an additional Bayer mask.
A generalization to spectral light fields in these instances is generally not straight-forward.
Two approaches have been studied to spectrally code light fields: again, either coding the angular or the spatial component.
While coding the angular component can easily be done for camera arrays by placing a spectral filter in front of each individual camera~\cite{Zhu2018}, it is challenging for MLA-based cameras.
Previous studies have placed a spectral mask in the main lens plane~\cite{Horstmeyer2009,Meng2012}, however, alignment of the mask with the camera sensor is virtually impossible to achieve: each spectral mask segment has to be imaged onto exactly one pixel since the pixels underneath each microlens code the angular component.
 Hence, when inevitably misaligned, the resulting coding is in fact not purely angular.
Furthermore, the misalignment of the MLA and the sensor cannot be corrected via interpolation since each microlens image is only sparsely sampled.
On the other hand, spatial coding in the case of camera arrays can be achieved by placing the same spectral mask in front of each camera sensor.
For MLA-based cameras in the unfocused design~\cite{Ng2005}, the same can be accomplished by coding the MLA~\cite{Ye2015}, resulting in a spatio-spectral coding which we elaborate shortly.
For completeness, the conventional Bayer sensor of an MLA-based light field camera also uses a spatio-angular color mask.
But the coding is not adapted to the light field geometry.
The sensor image is usually first demosaiced and then decoded to an RGB light field.
While light field camera-specific demosaicing methods exist~\cite{Yu2012,David2017}, the geometric properties of the light field are not explicitly taken into account.
Furthermore, this approach cannot be generalized to the multispectral case.
While it is feasible in the case of conventional cameras~\cite{Geelen2014}, the number of measurements per microlens would simply be too sparse for standard sensor-based demosaicing.
Hence, a more adapted spectral coding is needed.
We believe that for hand-held MLA-based light field cameras, only the coding of the MLA is truly practical as the coding naturally aligns with the discrete sampling of the light field by the MLA.
The light field decoding can be performed in complete analogy to the RGB case since every microlens image is fully sampled and can be aligned with the sensor via standard procedures~\cite{Dansereau2013}.

\textbf{Spatio-spectrally coded light fields:} Whereas the other coding schemes have been thoroughly discussed in the literature, coding of the MLA has attracted only little attention~\cite{Ye2015}.
While this might be due to a challenging hardware realisation, spectrally coded MLAs have become feasible using conventional lithographic methods and modern techniques such as micro-optic inkjet printing~\cite{Alaman2016,Cox2001}.

\begin{figure}[t]
	\centering
	\begin{tikzpicture}[>=latex,samples=100, scale=0.65]

	\tikzstyle{every node}=[font=\scriptsize]
	\pgfpointtransformed{\pgfpointxy{1}{1}}
	\pgfgetlastxy{\vx}{\vy}

	\begin{scope}[node distance=\vy and \vx]

		\pgfmathsetmacro{\H}{1.95}

		\pgfmathsetmacro{\posx}{-3.5}
		\pgfmathsetmacro{\posy}{0}

		\pgfmathsetmacro{\d}{0.5}

		\pgfmathsetmacro{\R}{\H}

		\pgfmathsetmacro{\rr}{8.5}

		\pgfmathsetmacro{\rl}{8.5}

		\pgfmathsetmacro{\rphi}{asin(\R / \rr)} 
		\pgfmathsetmacro{\lphi}{asin(\R / \rl)} 
		\pgfmathsetmacro{\dr}{\rr-\rr*cos(\rphi)} 
		\pgfmathsetmacro{\dl}{\rl-\rl*cos(\lphi)} 

		\pgfmathsetmacro{\posMLA}{0.97}
		\pgfmathsetmacro{\tiltMLA}{0} 
		\pgfmathsetmacro{\cosTilt}{cos(-\tiltMLA)} 
		\pgfmathsetmacro{\sinTilt}{sin(-\tiltMLA)} 
		\pgfmathsetmacro{\oR}{0.2}
		\pgfmathsetmacro{\oG}{-0.15}
		\pgfmathsetmacro{\dia}{0.52}
		\pgfmathsetmacro{\f}{1.45}
		\pgfmathsetmacro{\F}{-1*\posx + \posMLA}


		\pgfmathsetmacro{\MLd}{0.15}

		\pgfmathsetmacro{\MLR}{0.5*\dia}

		\pgfmathsetmacro{\MLrr}{0.5}

		\pgfmathsetmacro{\MLrl}{\MLrr}

		\pgfmathsetmacro{\MLrphi}{asin(\MLR / \MLrr)} 
		\pgfmathsetmacro{\MLlphi}{asin(\MLR / \MLrl)} 
		\pgfmathsetmacro{\MLdr}{\MLrr-\MLrr*cos(\MLrphi)} 
		\pgfmathsetmacro{\MLdl}{\MLrl-\MLrl*cos(\MLlphi)} 

		\pgfmathsetmacro{\posS}{\posMLA + \f}
		\pgfmathsetmacro{\w}{1.95*\H}

		\pgfmathsetmacro{\lim}{3}

		\draw[thin,fill=kit-blue, fill opacity=0.2] (\posx,-\R+\posy) -- (.5*\d-\dr+\posx,-\R+\posy) arc (-\rphi:\rphi:\rr) -- (-.5*\d+\dl+\posx,\R+\posy) arc (180-\lphi:180+\lphi:\rl) -- cycle;
		\path[name path=leftarc] (-.5*\d+\dl+\posx,\R+\posy) arc (180-\lphi:180+\lphi:\rl);
		\path[name path=rightarc] (.5*\d-\dr+\posx,-\R+\posy) arc (-\rphi:\rphi:\rr);
		\node[below, xshift=-0.8cm, yshift=3mm] at (\posx,-\H) {\tikzsize main lens};

		\draw[kit-blue, very thick, <-] (\posx - 1.5, 0) -> (\posS, 0);
		\draw[kit-blue, very thick, ->] (\posx, -\H-0.05) -> (\posx, \H + 0.5);
		\node[above] at (\posx - 0.63, \H - 0.05) {\tikzsize $(u,v)$};

		\draw[thick] (\posS,-.54*\w) -> (\posS,.57*\w);
		\node[below, xshift=5mm, yshift=3mm] at (\posS,-\H) {\tikzsize sensor};

		\draw[semithick, rotate around={\tiltMLA:(\posMLA, \oR)}] (\posMLA,-.475*\w) -> (\posMLA,.5*\w);

		\node[below, xshift=-0.55cm, yshift=3mm] at (\posMLA,-\H) {\tikzsize MLA};

		\pgfmathsetseed{330}

		\foreach \i in {-3,-2,...,3}{

			\pgfmathtruncatemacro{\j}{\i}

			\pgfmathsetmacro{\cx}{\posMLA + \oG*\sinTilt + \i*\dia*\sinTilt}
			\pgfmathsetmacro{\cy}{\oR + \oG*\cosTilt + \i*\dia*\cosTilt}

			\pgfmathsetmacro{\lambda}{(\posS - \posx)/(\cx - \posx)}
			\pgfmathsetmacro{\cxp}{\posx + \lambda*\cx - \lambda*\posx}
			\pgfmathsetmacro{\cyp}{\lambda*\cy}



			\pgfmathparse{rnd}
			\xdefinecolor{RandomColor}{hsb}{\pgfmathresult, 0.85, 0.95}

			\draw[color=kit-blue50, thin] (\posx, \posy) -> (\cx, \cy); 
			\draw[color=RandomColor, thin] (\cx, \cy) -> (\cxp, \cyp); 

			\draw[color=RandomColor, thin,fill=RandomColor, fill opacity=0.9] (\cx,-\MLR+\cy) -- (.5*\MLd-\MLdr+\cx,-\MLR+\cy) arc (-\MLrphi:\MLrphi:\MLrr) -- (-.5*\MLd+\MLdl+\cx,\MLR+\cy) arc (180-\MLlphi:180+\MLlphi:\MLrl) -- cycle;
			\path[name path=leftarc] (-.5*\MLd+\MLdl+\cx,\MLR+\cy) arc (180-\MLlphi:180+\MLlphi:\MLrl);
			\path[name path=rightarc] (.5*\MLd-\MLdr+\cx,-\MLR+\cy) arc (-\MLrphi:\MLrphi:\MLrr);

		}

		\path[\myarrowhead-\myarrowhead] (\posx, \H) edge[thin, shorten >=1pt, shorten <=1pt] node[fill=white, anchor=center, pos=0.5] {\tikzsize $I$} (\posMLA, \H);
		\path[\myarrowhead-\myarrowhead] (\posMLA, \H) edge[thin, shorten >=1pt, shorten <=1pt] node[fill=white, anchor=center, pos=0.5] {\tikzsize $f$} (\posS, \H);

		\node[above] at (\posMLA, \H - 0.05) {\tikzsize $(s,t)$};

	\end{scope}

\end{tikzpicture}
	\caption{Light field camera-model with spectrally coded MLA.}
	\label{fig:camera-model}
\end{figure}
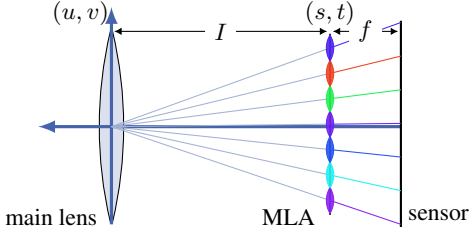

For light field cameras in the unfocused design, the sensor is placed at a distance from the MLA equal to the focal length $f$ of the microlenses as depicted in Figure~\ref{fig:camera-model}.
The MLA itself is placed in the focal plane with image distance $I$ of the main lens.
By spectrally coding the individual microlenses of the MLA, such that each microlens acts as a spectral bandpass filter, one obtains the spatio-spectrally coded light field
\begin{equation}
	\mathcal{L}^{*}[u, v, s, t, \lambda] = \mathcal{M}[s, t, \lambda] \cdot \mathcal{L}[u, v, s, t, \lambda]\,.
\end{equation}
Here, $\mytensor{M} \in \left\{0, 1\right\}^{S \times T \times \Lambda}$ denotes the binary coding mask.
Since only one filter is used in the imaging process at every spatial coordinate $(s, t)$,  $\mytensor{M}$ fulfills the one-hot constraint
\begin{equation}\label{eq:summation-constraint}
	\sum\nolimits_{\lambda = 1}^{\Lambda} \mathcal{M}[s, t, \lambda] = 1 \,.
\end{equation}
During the measurement, the coded light field is projected along the spectral dimension,
\begin{equation}
	\mathcal{L}^{*}_{\textrm{p}}[u, v, s, t] = \sum\nolimits_{\lambda = 1}^{\Lambda} \mathcal{L}^{*}[u, v, s, t, \lambda]\,.
\end{equation}
When the coding mask $\mytensor{M}$ is known, which can be achieved during calibration, the coded $\mytensor{L}^{*}$ can be obtained from its projection $\mytensor{L}^{*}_{\textrm{p}}$ since for every pixel only one spectral channel has a non-zero value in $\mytensor{L}^{*}$.
Therefore, we consider $\mytensor{L}^{*}$ and $\mytensor{L}^{*}_{\textrm{p}}$ to be equivalent in the following.

\section{Reconstruction from Coded Light Fields}

Traditionally, the full light field $\mytensor{L}$ is recovered from the coded measurement ${\mytensor{L}}^{*}$.
For example, the Bayer-coded sensor image is demosaiced before the light field is decoded from it.
When high compression ratios are involved, which is the case for coded light field camera designs and in particular when considering spectral light fields, the reconstruction can be done in the CS framework~\cite{Eldar2012:Compressed-sensing,Marwah2013,Vadathya2019:Unified-learning-based-framework} by maximizing the reconstructed signal's sparsity in some adapted frame or basis while being conform to the coded measurement.
From the fully recovered light field, the desired information is subsequently extracted.
We consider two CS-based reconstruction methods which we discuss in Section~\ref{sec:baseline-methods}.

\subsection{Proposed Principal Reconstruction}
\label{sec:principal-reconstruction}

Instead of reconstructing an intermediate full light field from the coded measurement, we propose to infer the desired properties from the coded light field directly.
That is, given the coded measurement $\mytensor{L}^{*}$, estimating the central view $\mathcal{I}_\textrm{c}[s, t, \lambda]$ and the corresponding disparity map $D_\textrm{c}[s, t]$ without the intermediate recovery of the full light field $\mytensor{L}$.
In this instance, the light field camera with a spectrally coded MLA can be interpreted as a single-shot spectral depth camera.
We use an encoder-decoder network with multiple decoder streams to decode the spectral central view and disparity from the coded light field.
In principle, the proposed method is not specific to the low dimensional reconstruction target---one could equally consider segmentation, saliency or reflection properties of the light field.
A schematic comparison of conventional and principal reconstruction, in the case of the used targets $\mytensor{I}_\textrm{c}$ and $\myvec{D}_\textrm{c}$, is shown in Figure~\ref{fig:comparison-traditional-principal}.

\subsection{Network Architecture}
\label{sec:network-architecture}

Recently, many deep learning-based light field applications have been studied.
In the case of disparity estimation, neural networks have significantly outperformed traditional methods~\cite{Ma2018,Shin2018:EPINET}.
Furthermore, neural networks have been applied to light field superresolution~\cite{Yeung2019,Yoon2017}, intrinsics estimation~\cite{Alperovich2018}, dense-from-sparse light field reconstruction~\cite{WingFungYeung2018}, classification~\cite{Wang2016a}, and more.
Commonly, light field-related deep learning approaches use the epipolar plane image (EPI)-volumes as the network's input~\cite{Alperovich2018,Heber2016,Heber2017,Shin2018:EPINET}.
An EPI-volume is a 3D section of the 4D light field, fixing one of the two angular coordinates.
Usually, the central angular coordinates are fixed, resulting in two geometrically perpendicular 3D EPI-volumes which are used as the input.
Alternatively, stereo-view pairs~\cite{Shi2019} or either the full or a sparse subsets of the light field are used~\cite{Ma2018,Peng2018,Yeung2019,Yoon2017}.

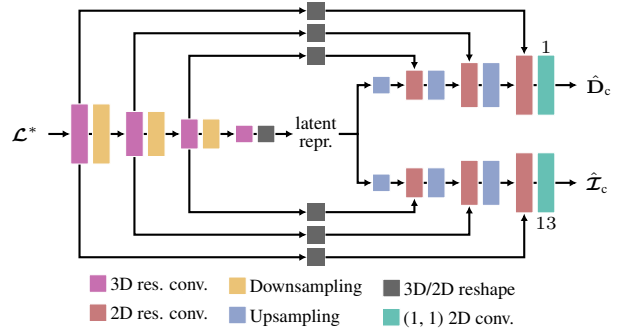
\begin{figure}[t]
	\centering
	\begin{tikzpicture}[node distance = 0.44cm]
	\tikzstyle{every node}=[font=\scriptsize]
	
	\newcommand{\NETBLOCKHEIGHT}{8mm}
	\newcommand{\NETBLOCKWIDTH}{1mm}
	\newcommand{\NETBLOCKSEP}{1.5mm}
	
	\tikzstyle{netblockmaster} = [draw, minimum width=\NETBLOCKWIDTH, minimum height=\NETBLOCKHEIGHT]
	\tikzstyle{netblockslave} = [netblockmaster, xshift=-\NETBLOCKSEP]
	\tikzstyle{legend} = [draw, minimum width=\NETBLOCKWIDTH, minimum height=\NETBLOCKWIDTH, xshift=-3mm]
	
	\tikzstyle{sizefull} = [minimum height=\NETBLOCKHEIGHT]
	\tikzstyle{sizethreeq} = [minimum height=0.75*\NETBLOCKHEIGHT]
	\tikzstyle{sizehalf} = [minimum height=0.5*\NETBLOCKHEIGHT]
	\tikzstyle{sizeq} = [minimum height=0.25*\NETBLOCKHEIGHT]

	\tikzstyle{labelabove} = [yshift=1mm]
	\tikzstyle{labelbelow} = [yshift=-1mm]
	\tikzstyle{legendlabel} = [xshift=-4mm, yshift=-0.3mm]
	
	\tikzstyle{downsample} = [white, fill=kit-blue, fill opacity=0.6]
	\tikzstyle{residual3d} = [white, fill=kit-purple, fill opacity=0.6]
	\tikzstyle{downsample} = [white, fill=kit-orange, fill opacity=0.6]
	\tikzstyle{reshape} = [white, fill=black, fill opacity=0.6]
	\tikzstyle{upsample} = [white, fill=kit-blue, fill opacity=0.6]
	\tikzstyle{residual2d} = [white, fill=kit-red, fill opacity=0.6]
	\tikzstyle{final} = [white, fill=kit-green, fill opacity=0.6]

	\node (input) {$\mytensor{L}^{*}$};
	\node[labelbelow] at (input.south) {};
	
	\node[right=of input, netblockmaster, residual3d, sizefull, xshift=-1.5mm] (e1a) {};
	\node[right of=e1a, netblockslave, downsample, sizefull] (e1b) {};
	
	\node[right of=e1b, netblockmaster, residual3d, sizethreeq] (e2a) {};
	\node[right of=e2a, netblockslave, downsample, sizethreeq] (e2b) {};
	
	\node[right of=e2b, netblockmaster, residual3d, sizehalf] (e3a) {};
	\node[right of=e3a, netblockslave, downsample, sizehalf] (e3b) {};
	
	\node[right of=e3b, netblockmaster, residual3d, sizeq] (e4a) {};
	\node[right of=e4a, netblockslave, reshape, sizeq] (e4b) {};

	\node[right=of e4b, xshift=-2mm] (latent) {\phantom{late}};
	\node[yshift=+1.5mm] at (latent) {latent};
	\node[yshift=-1.5mm] at (latent) {repr.};
	
	\node[right=of latent, yshift=0.65cm, netblockmaster, upsample, sizeq] (d11) {};
	
	\node[right of=d11, netblockmaster, residual2d, sizehalf] (d12a) {};
	\node[right of=d12a, netblockslave, upsample, sizehalf] (d12b) {};
	
	\node[right of=d12b, netblockmaster, residual2d, sizethreeq] (d13a) {};
	\node[right of=d13a, netblockslave, upsample, sizethreeq] (d13b) {};
	
	\node[right of=d13b, netblockmaster, residual2d, sizefull] (d14a) {};
	\node[right of=d14a, netblockslave, final, sizefull] (d14b) {};
	\node[labelabove] at (d14b.north) {$1$};
	
	\node[right=of d14b,xshift=-1.5mm] (outdisp) {$\hat{\myvec{D}}_{\textrm{c}}$};
	
	\node[right=of latent, yshift=-0.65cm, netblockmaster, upsample, sizeq] (d21) {};
	
	\node[right of=d21, netblockmaster, residual2d, sizehalf] (d22a) {};
	\node[right of=d22a, netblockslave, upsample, sizehalf] (d22b) {};
	
	\node[right of=d22b, netblockmaster, residual2d, sizethreeq] (d23a) {};
	\node[right of=d23a, netblockslave, upsample, sizethreeq] (d23b) {};
	
	\node[right of=d23b, netblockmaster, residual2d, sizefull] (d24a) {};
	\node[right of=d24a, netblockslave, final, sizefull] (d24b) {};
	\node[labelbelow, yshift=-0.25mm] at (d24b.south) {$13$};
	
	\node[right=of d24b,xshift=-1.5mm] (outcentral) {$\hat{\mytensor{I}}_{\textrm{c}}$};
	
	\draw[-\myarrowhead, thick] (input)  -- (e1a);
	\draw[thick] (e1a)  -- (e1b);
	\draw[-\myarrowhead, thick] (e1b)  -- (e2a);
	\draw[thick] (e2a)  -- (e2b);
	\draw[-\myarrowhead, thick] (e2b)  -- (e3a);
	\draw[thick] (e3a)  -- (e3b);
	\draw[-\myarrowhead, thick] (e3b)  -- (e4a);
	\draw[thick] (e4a)  -- (e4b);
	\draw[-\myarrowhead, thick] (e4b)  -- (latent);
	
	\draw[-\myarrowhead, thick] (latent.east) -- +(0.25cm, 0) |- (d11);
	\draw[-\myarrowhead, thick] (d11)  -- (d12a);
	\draw[thick] (d12a)  -- (d12b);
	\draw[-\myarrowhead, thick] (d12b)  -- (d13a);
	\draw[thick] (d13a)  -- (d13b);
	\draw[-\myarrowhead, thick] (d13b)  -- (d14a);
	\draw[thick] (d14a)  -- (d14b);
	\draw[-\myarrowhead, thick] (d14b)  -- (outdisp);
	\draw[-\myarrowhead, thick] (latent.east) -- +(0.25cm, 0) |- (d21);
	\draw[-\myarrowhead, thick] (d21)  -- (d22a);
	\draw[thick] (d22a)  -- (d22b);
	\draw[-\myarrowhead, thick] (d22b)  -- (d23a);
	\draw[thick] (d23a)  -- (d23b);
	\draw[-\myarrowhead, thick] (d23b)  -- (d24a);
	\draw[thick] (d24a)  -- (d24b);
	\draw[-\myarrowhead, thick] (d24b)  -- (outcentral);
	
	\node[above of=latent, reshape, yshift=12mm] (r11) {};
	\draw[-\myarrowhead, thick] (e1a)  |- (r11);
	\draw[-\myarrowhead, thick] (r11) -| (d14a);
	
	\node[above of=latent, reshape, yshift=9mm] (r12) {};
	\draw[-\myarrowhead, thick] (e2a)  |- (r12);
	\draw[-\myarrowhead, thick] (r12) -| (d13a);
	
	\node[above of=latent, reshape, yshift=6mm] (r13) {};
	\draw[-\myarrowhead, thick] (e3a)  |- (r13);
	\draw[-\myarrowhead, thick] (r13) -| (d12a);
	
	\node[below of=latent, reshape, yshift=-12mm] (r21) {};
	\draw[-\myarrowhead, thick] (e1a)  |- (r21);
	\draw[-\myarrowhead, thick] (r21) -| (d24a);
	
	\node[below of=latent, reshape, yshift=-9mm] (r22) {};
	\draw[-\myarrowhead, thick] (e2a)  |- (r22);
	\draw[-\myarrowhead, thick] (r22) -| (d23a);
	
	\node[below of=latent, reshape, yshift=-6mm] (r23) {};
	\draw[-\myarrowhead, thick] (e3a)  |- (r23);
	\draw[-\myarrowhead, thick] (r23) -| (d22a);
	
	\node[below of=latent, xshift=-2.6cm, yshift=-1.55cm, legend, residual3d] (l1) {};
	\node[right of=l1, anchor=west, legendlabel] (l1label) {3D res. conv.\vphantom{p}};

	\node[right = of l1label, legend, downsample] (l2) {};
	\node[right of=l2, anchor=west, legendlabel] (l2label) {Downsampling\vphantom{p}};

	\node[right = of l2label, legend, reshape] (l3) {};
	\node[right of=l3, anchor=west, legendlabel] (l3label) {3D/2D reshape\vphantom{p}};

	\node[below = of l1, legend, residual2d, xshift=3mm, yshift=3mm] (l4) {};
	\node[right of=l4, anchor=west, legendlabel] (l4label) {2D res. conv.\vphantom{p}};

	\node[right = of l4label, legend, upsample] (l5) {};
	\node[right of=l5, anchor=west, legendlabel] (l5label) {Upsampling\vphantom{p}};

	\node[right = of l5label, legend, final, xshift=2.9mm] (l6) {};
	\node[right of=l6, anchor=west, legendlabel] (l6label) {(1, 1) 2D conv.\vphantom{p}};

\end{tikzpicture}\vspace{-1.5mm}
	\caption{Overview of the dual-task encoder-decoder network.
	Note the encoder side uses 3D while the decoder
	uses 2D convolutions.}
	\label{fig:network}
\end{figure}

Since in our application we are dealing with coded light fields, and hence only few measurements per subaperture, we use the full (coded, non-projected) light field as its input.
We stack the coded subapertures along a single axis and use 3D convolutions to extract spatio-angular features.

For its versatility, we use a dual-stream U-net architecture~\cite{Ronneberger2015:Unet}, \ie an encoder-decoder network with skip connections.
The encoder maps the coded light field to a well adapted, low dimensional latent space.
Then, using two jointly trained decoder paths, the central view and the disparity map are decoded from the latent representation.
As input, the non-projected $\mytensor{L}^*$ is used.
This way, the coding mask is implicitly available to the network.
We stack the subapertures of the light field along a new axis, resulting in an input of shape $(S, T, U\!\cdot\!V, \Lambda)$.
The encoder path is then built upon 3D residual convolution blocks.
Spatio-angular downsampling is performed via strided 3D convolution.
The decoder paths are built using 2D residual and transposed convolutions because the disparity and central view do not have an angular dependence.
Therefore, 3D to 2D reshapes are necessary when connecting the encoder and decoders.
This way, our approach combines a 3D U-net encoder~\cite{Cicek2016:Unet3D} with dual stream 2D U-net decoders~\cite{Ronneberger2015:Unet}.
The proposed network architecture is depicted in Figure~\ref{fig:network}.
More details on the proposed architecture and convolution blocks can be found in the supplementary material.

\textbf{Related works:}
Two recent publications resemble our approach.
Vadathya~\etal~\cite{Vadathya2019:Unified-learning-based-framework} propose a general framework for light field reconstruction from coded projections.
In fact, similar to our approach, they estimate an intermediate central view and disparity field directly from the coded measurement.
However, there are some crucial differences to our work.
First, the estimated central view and disparity are used only intermediately to reconstruct the full light field from the coded measurement.
In fact, the estimation networks cannot be trained without the full light field reconstruction as the network design and loss function are based on the full reconstruction.
This directly opposes our approach.
However, due to the self-supervised formulation, their approach can be trained using real-world data and does not require synthetic disparity ground truth data, which is useful in practice.
Second, the considered coding schemes include angular integration and are only valid for attenuation mask-based compressive light field imagers.
This does not include MLA-multiplexed coded light fields which we consider.
And third, their approach, as many compressive light field approaches before, does not consider the color or spectral domain of the light field.
The impact of demosaicing in compressive light field imaging remains unclear and has yet to be investigated.

Furthermore, a new spectral depth camera, based on an end-to-end optimized free-form diffractive lens, has recently been proposed by Baek~\etal~\cite{Baek2020:End-to-end-hyperspectral-depth}.
The reconstruction targets are almost the same as ours---namely a spectral image and its aligned depth map.
The crucial difference to our approach is that we explicitly use the angular component of the incoming signal instead of a phase modulation.
This makes it possible to directly use the angular information in our decoding scheme which may be beneficial for the disparity estimation.
On the other hand, the approach by Baek~\etal is quite general and considers higher spatial and spectral resolutions than ours.
However, due to the joint hard- and software design, a fair quantitative comparison of the different imaging setups is non-trivial or even impossible.

\subsection{Training Strategies}
\label{sec:training-strategies}

\textbf{Multi-task training:}
Using the proposed network architecture, the training is inherently a multi-task problem.
The naive multi-task approach is to use a weighted sum of the individual task losses $L_i$ as the overall training loss
\begin{equation}\label{eq:mt-main-loss}
	L = \sum\nolimits_{i=1}^{N} w_{i}L_{i} + L_{\textrm{reg}} \,,
\end{equation}
in the case of $N$ tasks.
Here, $w_i > 0$ are the task weights and $L_{\textrm{reg}}$ is a task-independent regularization term (such as weight decay) which we neglect in the following.
The challenge in this straight-forward approach is to find suitable task weights $w_i$ which is time- and resource-intensive.
Furthermore, it may not be possible to find optimal \emph{static} weights.
During training, the gradient of the loss with respect to the network weights is calculated.
For each task, these weights contain some shared across all tasks (in our case the encoder weights), as well as task-specific weights (the individual decoder weights).
The shared weights are hence updated based on the gradients from all tasks which can be problematic:
the gradients from the different tasks may be on different scales leading to an imbalance during the weight update.
Furthermore, the scales of the gradients may change differently during training.
Finally, the tasks may also be of different complexity, leading to different convergence speeds which further enhances the task imbalance.
To overcome these difficulties, some approaches to \emph{dynamically} update the task weights $w_i$ during training have been recently discussed.
Here, we consider the Multi-Task with Uncertainty (MTU) approach~\cite{Kendall2018:MultiTaskUncertainty}, as well as the GradNorm method~\cite{Chen2018:GradNorm}, both of which have shown good results in the context of computer vision.
In both cases, the loss weights $w_i$ are considered adaptive or even trainable themselves.
While the MTU approach proposes a new loss function obtained from the maximum log-likelihood of the model taking into account the individual task estimation uncertainties, GradNorm explicitly considers the task gradient norms and convergence speeds.
For further technical details, we refer to the original literature.
When the tasks compete, it may be necessary to adapt methods from multi-objective learning~\cite{Sener2018:Multi-objective-optimization,Lin2019:pareto-multi-task-learning}.
In the explored case of central view and disparity estimation however, we find that the used multi-task approaches perform on par or better than the corresponding single-task networks.
Therefore, we did not explore multi-objective approaches.

\textbf{Auxiliary loss training:}
In computer vision, often the mean squared error (MSE) or mean absolute error (MAE) is used as a single-task loss function in regression tasks.
The use of the MSE as the primary optimization and evaluation metric is well justified since it corresponds to the energy of the reconstruction error.
Furthermore, the MSE is convex and continuously differentiable.
However, both MSE and MAE are only evaluated pixel-wise and subsequently averaged.
In particular, neither spatial nor spectral correlations are considered.
Often, however, secondary quality measures are of key interest. \Eg the structural similarity~(SSIM)~\cite{Wang2004} to asses the spatial reconstruction quality, or the spectral angle and the spectral information divergence~\cite{Chang2000:spectral-information-divergence} to evaluate the spectral reconstruction quality.
To utilize these secondary metrics during optimization, ignoring the multi-task scenario for now, one can use the loss function
\begin{equation}
	L = L_{\textrm{main}} + \sum\nolimits_{j = 1}^{N_{\textrm{aux}}} w_{\textrm{aux}, j}L_{\textrm{aux}, j}
\end{equation}
using $N_{\textrm{aux}}$ auxiliary loss functions $L_{\textrm{aux}, j}$ to support the main loss $L_{\textrm{main}}$.
In principle, the problems with this naive approach are similar to those in multi-task learning.
Namely, the manual (static) choice of the auxiliary loss weights and the possibly different scales of the loss gradients.
Furthermore, the additional losses may even be adversarial to the main loss, leading to cancelling gradients, and the overall loss landscape may suffer from high curvature~\cite{Yu2020:Gradient-surgery}.
To mitigate this, the use of gradient similarity has been proposed~\cite{Du2018:GradientSimilarity}.
Here, for each mini-batch during the training, the auxiliary loss weights are calculated using the truncated cosine similarity between the main and the auxiliary task gradients $\myvec{G}_{\textrm{main}}$ and $\myvec{G}_{\textrm{aux,j}}$.
While this resolves the issue concerning adversarial auxiliary losses and cancelling gradients, the problem of different gradient norms persists.
Furthermore, since the calculation is performed for each mini-batch individually, the result may be quite noisy.
To overcome these limitations we propose a modification which we call Normalized Gradient Similarity~(NormGradSim).
We extend the single-task loss, using $N_{\textrm{aux}}$ auxiliary losses $L_{\textrm{aux}, j}$, to
\begin{equation}
	L =
	\Big(L_{\textrm{main}} + \sum\nolimits_{j = 1}^{N_{\textrm{aux}}} \alpha_{j}\beta_{j}L_{\textrm{aux}, j}\Big) \Big/ \Big(1 + \sum\nolimits_{j = 1}^{N_{\textrm{aux}}} \alpha_j\Big)
\end{equation}
 with dynamic weights $\alpha_j, \beta_j > 0$ trained using the losses
\begin{align}
	L_{\alpha}
	&= \sum\nolimits_j \Bigg\lvert
		\alpha_j - \max \left\{0, \frac{\langle \myvec{G}_{\textrm{main}}, \myvec{G}_{\textrm{aux}, j} \rangle}{\lVert \myvec{G}_{\textrm{main}} \rVert\cdot\lVert \myvec{G}_{\textrm{aux}, j} \rVert}\right\}
	\Bigg\rvert \label{eq:loss-alpha}\,,\\
L_{\beta} &= \sum\nolimits_j \Big\lvert \beta_j \cdot \lVert \myvec{G}_{\textrm{aux}, j} \rVert - \lVert \myvec{G}_{\textrm{main}} \rVert  \Big\rvert  \label{eq:loss-beta}\,.
\end{align}
While $\alpha_j$ is used to weigh the auxiliary loss according to its gradient similarity, $\beta_j$ is used to bring the gradient of the auxiliary loss to the same scale as the main loss.
The normalization by $(1 + \sum_j \alpha_j)$ keeps the resulting gradient of the total loss $L$ on the same scale as the original main loss $L_{\textrm{main}}$.
This has the advantage that NormGradSim can be used as a drop-in replacement to the single loss training (without having to adapt hyperparameters such as the optimizer's learning rate) or combined with multi-task approaches such as MTU and GradNorm, as the scale is unchanged.

Combining the multi-task scenario with the proposed NormGradSim, we obtain the final overall loss function
\begin{equation}
	L = \sum_{i=1}^{N} w_{i} \Big( L_{\textrm{main}}^{(i)} + \sum_{j=1}^{N^{(i)}_\textrm{aux}} \alpha_j^{(i)}\beta_j^{(i)}\,L_{\textrm{aux},j}^{(i)} \Big) \Big/ \Big(1 + \sum_{j=1}^{N^{(i)}_\textrm{aux}} \alpha_j^{(i)} \Big)
\end{equation}
in combination with the additional loss functions $L_{\alpha}, L_{\beta}$.

\section{Experiment Details}
\label{sec:experiment-details}

\subsection{Dataset}
\label{sec:dataset}
For the supervised training, as well as the quantitative evaluation of the disparity estimation, synthetic spectral light field data is used.
This is common since there are no suitable depth reference measurement techniques available.
To further investigate the results, real-world spectral light field data is also evaluated.
Using fully sampled spectral light fields and simulating the spatio-spectral coding, the reconstruction of the central view can be quantitatively evaluated.

\textbf{Synthetic dataset:}
For synthetic spectral light fields we use the recently proposed dataset by Schambach and Heizmann~\cite{Schambach2020:Multispectral-lf-dataset}.
The dataset contains 500 spectral light fields of shape $(11, 11, 512, 512, 13)$.
For training, test, and validation, the dataset is split 400:50:50 and patched into small light field patches of shape $(9, 9, 36, 36, 13)$.
Further, the dataset provides hand-crafted spectral light fields of shape $(11, 11, 512, 512, 13)$.
We will use these dataset challenges to evaluate all methods also for full-sized prediction.
The angular resolution of the synthetic light fields is reduced to $(9 ,9)$ which is common in the light field community.

\textbf{Real-world dataset:}
To our knowledge, the only public spectral light field dataset was made available by Xiong~\etal~\cite{Xiong2017}.
This dataset consists of three spectral light fields (\textit{Boards, Toys, and Fruits}) of shape $(11, 11, 360, 270, 25)$ captured using a gantry-mounted 2D spectrometer.
However, the objects used in the dataset are not all textured which is not ideal for the evaluation of the disparity estimation.

Therefore, we created a spectral light field dataset using a custom-built spectral light field camera.
The camera consists of a Lytro Illum light field camera and a custom-built housing enclosing a filter wheel.
The wheel holds 13 spectral filters ranging from \SI{400}{\nano\meter} to \SI{700}{\nano\meter} with \SI{25}{\nano\meter} filter width.
In this configuration, the light fields are spectrally sampled in complete analogy to the used synthetic dataset.
To step the filter wheel automatically, the wheel is flange-mounted on a stepper motor.
The camera and the stepper motor are synchronously controlled by a Raspberry Pi.
Since the Lytro Illum uses a Bayer pattern RGB sensor, the spectral calibration of the camera is rather challenging.
We present our calibration procedure in the supplementary material.

\begin{figure}
	\centering\tikzsize
	\begin{minipage}[t]{0.32\linewidth}
		\vskip 0pt
		\centering
		\textit{Diavolo\vphantom{p}}\\[-0.25mm]
		\includegraphics[width=0.97\linewidth]{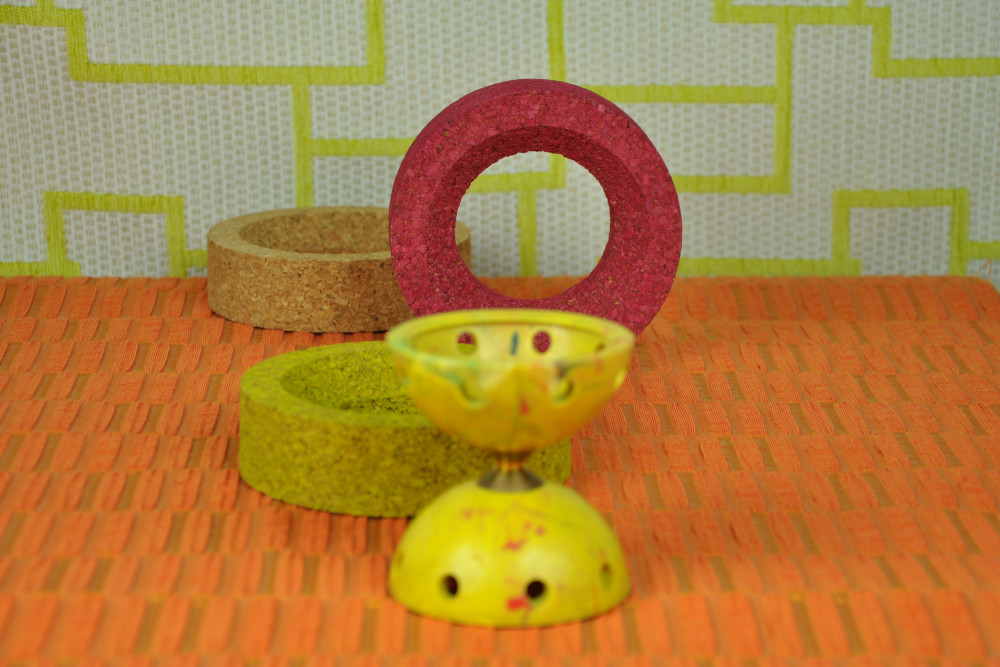}
	\end{minipage}
	\hfill
	\begin{minipage}[t]{0.32\linewidth}
		\vskip 0pt
		\centering
		\textit{Floral\vphantom{p}}\\[-0.25mm]
		\includegraphics[width=0.97\linewidth]{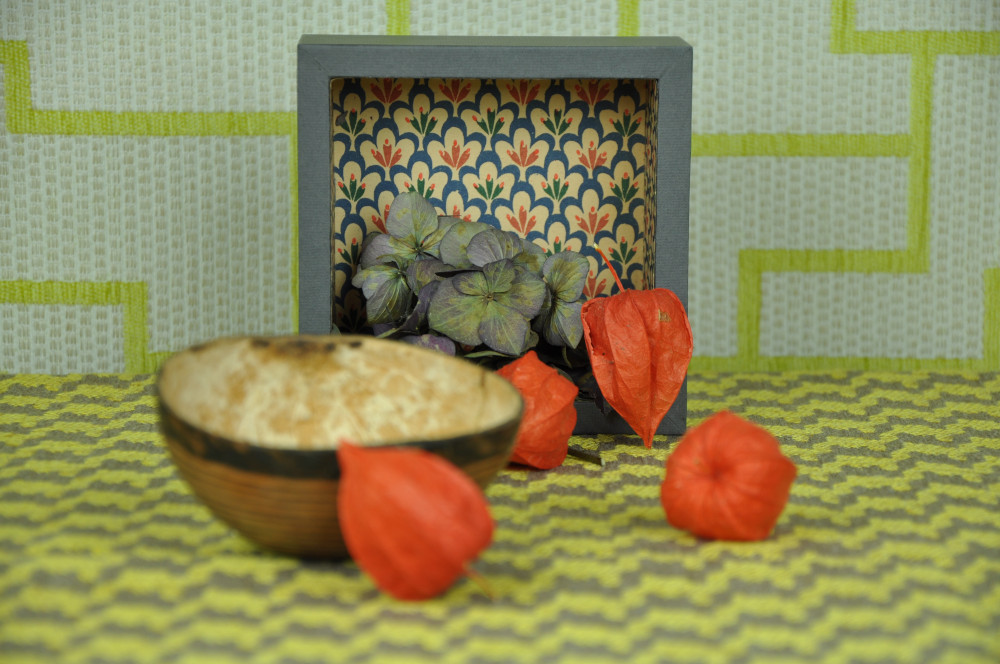}
	\end{minipage}
	\hfill
	\begin{minipage}[t]{0.32\linewidth}
		\vskip 0pt
		\centering
		\textit{Wagons\vphantom{p}}\\[-0.25mm]
		\includegraphics[width=0.97\linewidth]{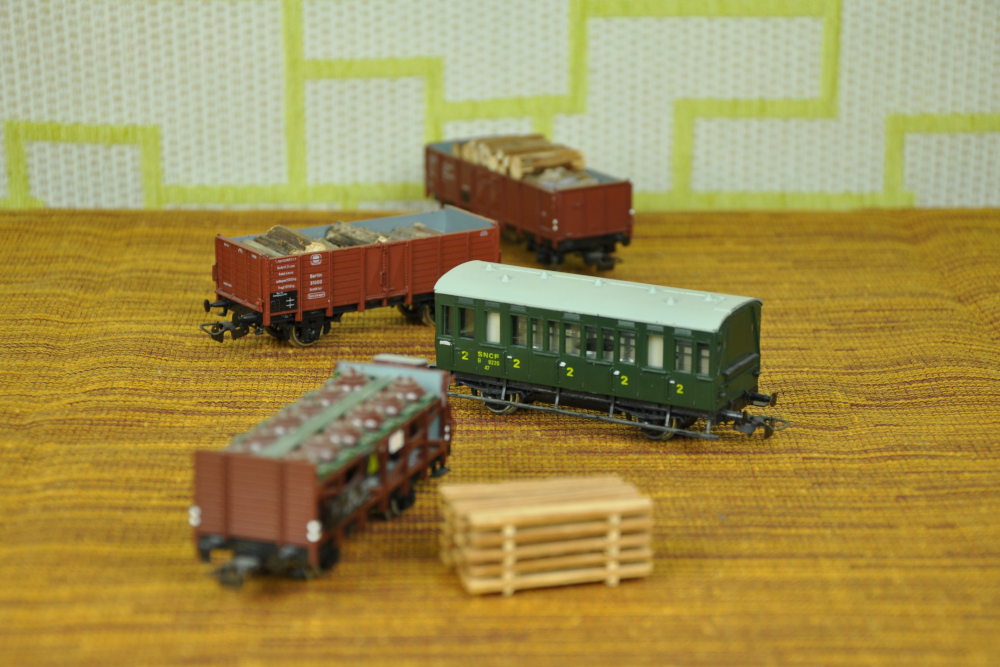}
	\end{minipage}\\[0.5mm]
	\caption{Dataset scenes captured with an RGB DSLR camera.}
	\label{fig:dataset-scenes}
	\vspace{-3mm}
\end{figure}

In total, we captured three different scenes, each using four different camera settings.
To reduce the influence of the angular dependence of the interference filters, and to increase the imaged disparity range, the scenes are captured using the comparably large \SI{150}{\milli\meter} and \SI{250}{\milli\meter} main lens focal length equivalent.
For each focal length setting, the scenes are imaged using two focus settings.
The captured scenes were created with spectral variance and disparity estimation in mind.
Hence, we used highly textured and colorful backgrounds and scene objects, and placed the objects as stretched out as possible to obtain a diverse disparity distribution.
An RGB view of the scenes is given in Figure~\ref{fig:dataset-scenes}.
In the used large focal length setting, the $f$-number matching of the Lytro Illum camera is suboptimal, leading to a smaller angular resolution than with shorter focal lengths.
We crop the angular resolution of the captured light fields to $(9, 9)$.
Spatially, the light fields are cropped to $(400, 400)$ to be compatible with the network's downsampling.

We will use spectral light fields from both datasets for the evaluation.
For compatibility, we downsampled the dataset by Xiong~\etal~\cite{Xiong2017} to 13 spectral channels and cropped the light fields to shape $(9, 9, 256, 360, 13)$.
As is common with real-world light field datasets, both datasets do not contain a depth or disparity reference.
Hence, they can not be used to quantitatively evaluate disparity estimation performance, however a qualitative comparison is of course possible.

\subsection{Baseline Methods}
\label{sec:baseline-methods}

Since a comparable approach to ours has not yet been considered, no baseline exists for comparison.
As we have elaborated, previous works on reconstruction from coded light fields cannot be applied to the used spatio-spectral light field coding.
Hence, we investigate several multi-task training strategies and compare them to their single-task variants.
Moreover, we compare the proposed approach with a CS reconstruction and subsequent disparity estimation.
We investigate two CS approaches to reconstruct the full light field from the coded measurement, one based on the 5D-DCT and one dictionary learning approach.
While full light field reconstruction is a more demanding task than the reconstruction of the central view alone, the full light field is needed for the disparity estimation.
We consider a well-established CS reconstruction based on the convex relaxation of the $l_0$-norm minimization.
For the 5D-DCT-based reconstruction, we use the OWL-QN variant of the L-BFGS optimizer~\cite{Byrd:1995:LBFGS,Andrew:2007:OWL-QN-LFBGS}.
In the case of the dictionary learning approach, we employ sparse coding using the Fast Iterative Shrinkage/Thresholding Alrogithm~\cite{Beck2009:FISTA} combined with a mini-batch SGD minimization of the reconstruction error.
Details on the investigated CS approaches can be found in the supplementary material.
For a comparison of the estimated disparity, we use the state-of-the-art EPINET~\cite{Shin2018:EPINET} using the reconstructed \emph{uncoded} light fields as input.
EPINET is one of the top-performing methods in the HCI~\cite{Honauer2016} for which a publication and reference implementation is available.

\subsection{Training Details}
\label{sec:training-details}

All learning-based methods were implemented using the LFCNN framework~\cite{Schambach2020:Multispectral-lf-dataset} and trained using the same training dataset.
During training, a random patch with spatial shape of $(32, 32)$ is cropped from the input light field.
For the light field coding, a random binary coding mask $\mytensor{M}$, satisfying the constraint~\eqref{eq:summation-constraint}, is drawn independently for each light field in the mini-batch.
This encourages the network to learn a mask-independent latent representation of the coded light field.
This has the advantage that coded light fields of different input shapes and different coding masks can be reconstructed by the network.
The trained network can hence be used for full-sized light field inference without modification.
In principle, using a regular coding mask is also possible.
But according to CS, random measurement matrices have a large incoherence with any orthonormal basis~\cite{Candes2006:Robust,Candes2007:Sparsity} providing certain guarantees on the reconstruction~\cite{Eldar2012:Compressed-sensing}.
For testing, the random seed of all random operations is fixed throughout all experiments, guaranteeing that the evaluation is always performed on the exact same coded data.
As the main loss, for both the central view as well as the disparity estimation, the Huber loss~\cite{Huber1992:Huber-loss} is used.
In the case of the auxiliary loss approaches, we employ an SSIM-based loss and a spectral-similarity loss for the central view task, and total variation-bases smoothness and normal similarity losses~\cite{Hu2019:Revisiting-single-image-depth} for the disparity task.
The EPINET was trained as specified in the original paper~\cite{Shin2018:EPINET}, for which the (uncoded) training dataset was converted to RGB.
Further details on the used training parameters, task weights and loss functions are presented in the supplementary material.

\begin{table}
	\centering
	\scriptsize
	\setlength{\tabcolsep}{2.25pt}
	\caption{Test dataset comparison of the Compressed Sensing (CS), single-task (ST), multi-task (MT), and auxiliary-loss (AL) methods.
	Proposed methods in blue.
	Best values in bold.
	The EPINET estimation from uncoded data is excluded from the comparison.}
	\vspace{2mm}
	\label{tab:performance-comparison}
	\begin{tabular}{lrrrrrr}
\toprule
                                                \multirow{3}{*}{\vspace{3mm}\thead{Method}}&\multicolumn{3}{c}{\thead{Central View}}&\multicolumn{3}{c}{\thead{Disparity}} \\ \cmidrule(lr){2-4} \cmidrule(lr){5-7} & \hspace{-4mm}\thead{PSNR $\!/\!$ dB} &          \thead{SSIM} & \thead{SA $\!/\!$ $^\circ$} &  \thead{MAE $\!/\!$ px} & \thead{MSE $\!/\!$ px$^2$} & \thead{BP07 $\!/\!$ \%} \\
\midrule
\textcolor{mygray}{EPINET~\cite{Shin2018:EPINET} (uncoded) } & \textcolor{mygray}{-} & \textcolor{mygray}{-} & \textcolor{mygray}{-} & \textcolor{mygray}{\num{0.0626}} & \textcolor{mygray}{\num{0.0881}} & \textcolor{mygray}{\num{6.28}}\\
                                    CS 5D-DCT + EPINET &                          \num{29.34} &            \num{0.77} &                  \num{7.65} &            \num{0.4875} &               \num{1.3671} &             \num{41.65} \\
                                     CS Dict. + EPINET &                          \num{27.86} &            \num{0.74} &                  \num{7.25} &            \num{0.2939} &               \num{0.6050} &             \num{32.34} \\[1mm]
 ST Central view &               {\bfseries\num{32.68}} &            \num{0.92} &                  \num{5.39} &                       - &                          - &                       - \\
                                          ST Disparity &                                    - &                     - &                           - &            \num{0.0679} &               \num{0.0607} &             \num{14.23} \\[1mm]
 MT Naive (baseline) &                          \num{27.70} &            \num{0.85} &                  \num{8.85} &            \num{0.0697} &               \num{0.0626} &             \num{14.92} \\
                  MT GradNorm~\cite{Chen2018:GradNorm} &                          \num{31.95} &            \num{0.92} &                  \num{5.80} &            \num{0.0815} &               \num{0.0692} &             \num{19.07} \\
MT Uncertainty~\cite{Kendall2018:MultiTaskUncertainty} &                          \num{31.18} &            \num{0.91} &                  \num{6.27} &            \num{0.0665} &               \num{0.0594} &             \num{13.87} \\
           AL GradSim~\cite{Du2018:GradientSimilarity} &                          \num{29.44} &            \num{0.93} &                  \num{6.73} &            \num{0.0830} &               \num{0.0721} &             \num{18.28} \\
                  \textcolor{MYDETAIL}{AL NormGradSim} &                          \num{28.52} &            \num{0.89} &                  \num{7.77} &            \num{0.0660} &               \num{0.0569} &             \num{14.04} \\[1mm]
 \textcolor{MYDETAIL}{MTU + AL} &                          \num{31.94} & {\bfseries\num{0.94}} &       {\bfseries\num{5.36}} & {\bfseries\num{0.0656}} &    {\bfseries\num{0.0567}} &  {\bfseries\num{13.58}} \\
\bottomrule
\end{tabular}

	\vspace{-2mm}
\end{table}

\section{Results}
\label{sec:results}

A comparison of all considered methods evaluated on the test dataset is given in Table~\ref{tab:performance-comparison}.
As evaluation metrics for the reconstructed central view we use the Peak Signal-to-Noise Ratio~(PSNR), the Structural Similarity Index Metric (SSIM)~\cite{Wang2004}, the Spectral Angle (SA), and the Spectral Information Divergence (SID)~\cite{Chang2000:spectral-information-divergence}.
For the disparity, we evaluate the Mean Absolute Error (MAE), the Mean Squared Error (MSE), as well as the BadPix07 (BP07) metric~\cite{Honauer2016}.
Regarding the CS methods we are surprised to see the straight-forward 5D-DCT approach outperform the dictionary learning one in terms of the overall reconstruction PSNR.
This might be because the DCT-based approach uses the full light field during the sparse coding while the light fields are patched for the dictionary-based method to match the light field atoms' shape.
However, in both cases we observe a drastic loss of quality of the disparity estimated from the reconstructed light field as compared to the EPINET prediction from uncoded light fields.
Even though the light field quality of the dictionary method is worse than the DCT-based one, the estimated disparity is much better.
This is likely because the dictionary takes into account the light field geometry, which the DCT-based approach does not.
For the succeeding disparity estimation, the improved light field geometry is of great importance.
Therefore, we only consider the dictionary-based approach in the following.

\begin{figure}
	\centering
	\tiny
	\begin{minipage}[t]{0.615\linewidth}
		\centering
		Central View\\[-1.25mm]
		\rule{0.96\textwidth}{0.5pt}
	\end{minipage}
	\begin{minipage}[t]{0.2\linewidth}
		\centering
		Disparity\\[-1.25mm]
		\rule{0.95\textwidth}{0.5pt}
	\end{minipage}
	\hfill
	\begin{minipage}[t]{0.15\linewidth}
	\end{minipage}\\[0.5mm]
	\includegraphics[scale=0.5]{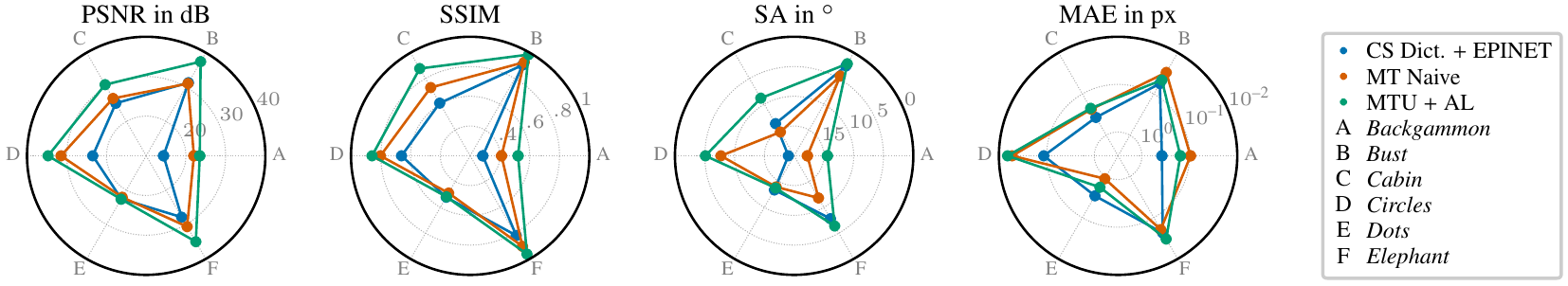}
	\caption{Performance comparison of the synthetic dataset challenges~\cite{Schambach2020:Multispectral-lf-dataset}.
	Outer values are better.}
	\label{fig:eval-challenges-radar}
\end{figure}

In the case of the considered deep learning approaches, we observe that the naive multi task approach shows a significant drop in performance compared to its single-task variants, most severely in the case of the central view estimation.
The two multi-task approaches, MTU and GradNorm, mitigate this loss in performance.
However, GradNorm seems not to be able to resolve the task imbalance, as the disparity estimation performs worse than the naive multi-task approach, even though we have performed a brute force optimization of the method's task imbalance hyperparameter.
In our application, the MTU approach yields a more balanced result than GradNorm, in contrast to the original findings~\cite{Chen2018:GradNorm}.
Considering the auxiliary loss methods, the standard gradient similarity approach~\cite{Du2018:GradientSimilarity} does improve the primary and secondary metrics in the case of the central view reconstruction.
However, the estimated disparity is worse than the MT baseline, likely due to an imbalance of the auxiliary task gradient norms.
Our proposed NormGradSim approach, using a normalized gradient similarity, resolves this issue, leading to an improvement of all considered primary and secondary metrics for both the central view as well as the disparity estimation when compared to the naive MT baseline.
Finally, combining the NormGradSim approach with the MTU method, the performance is further enhanced.
The result is only slightly outperformed by the single-task central view reconstruction in terms of the PSNR.
An evaluation of the methods using the synthetic dataset challenges and full-sized prediction is shown for all scenes in Figure~\ref{fig:eval-challenges-radar} and for the \emph{Elephant} scene in Figure~\ref{fig:eval-elephant}.
Visualizations of the remaining challenges can be found in the supplementary material.
The proposed combined approach outperforms the other methods in almost all considered metrics and scenes.
In particular, a significant improvement over the naive MT approach in the reconstructed central view can be observed.
Again, it can be observed that the disparity estimation using EPINET from the CS-reconstructed light fields shows severe artifacts.

\begin{figure}
	\centering
	\includegraphics[scale=0.5]{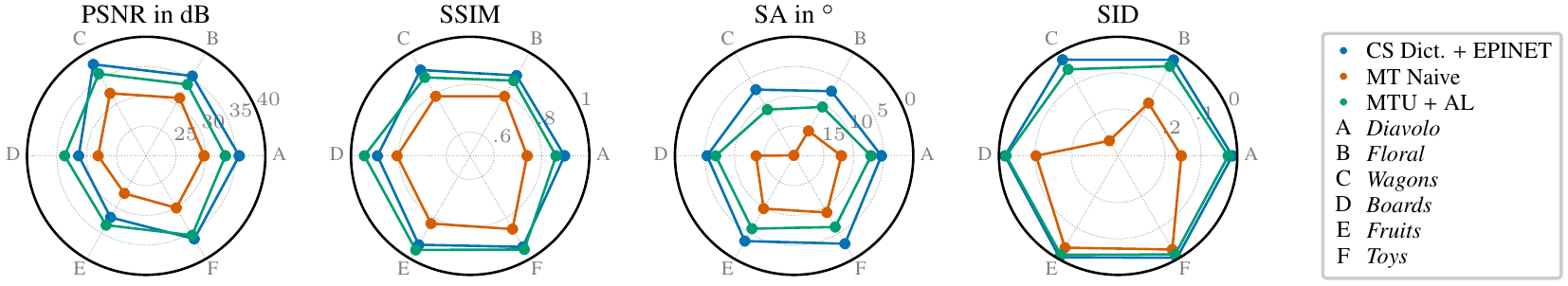}
	\caption{Performance comparison of the central view reconstruction from real-world datasets.
	For the proposed dataset, the scenes (A-C) with \SI{150}{mm} focal equivalent are used.
	The scenes D-F are from the dataset by Xiong~\etal~\cite{Xiong2017}.
	Outer values are better.
	}
	\label{fig:eval-real-data-radar}
	\vspace{-1mm}
\end{figure}

Regarding the real-world data, the results are a bit surprising.
Here, the CS-based method slightly outperforms the deep learning approaches in the case of the estimated central view, as depicted in Figure~\ref{fig:eval-real-data-radar} as well as Figure~\ref{fig:predict-real-diavolo}.
While the result visually appears quite blurry, the overall central view quality is higher than that of the proposed method.
However, the estimated disparity shows some significant artifacts, again supporting our claim that light field tasks suffer when applied to reconstructed light fields due to estimation noise and artifacts in the light field geometry.
On the other hand, the disparity estimated using the proposed approach also shows some significant artifacts, in particular in the background.
Therefore, also for the proposed method, a performance gap between synthetic and real-world data can be observed.
This issue, which is present in many light field deep learning approaches trained with synthetic data, remains an open quest in the light field community and needs to be addressed in future investigations.
Finally, note that the full-sized inference time of the proposed method is about \SI{8}{\second} per light field using an Nvidia Tesla V100 and can even be performed on mid-level 8\,GB gaming GPUs while the CS reconstruction takes more than 3\,h per light field using about 120\,GB of a 40-core computing node with 192\,GB RAM.

\begin{figure*}
	\centering
	\includegraphics[scale=0.96]{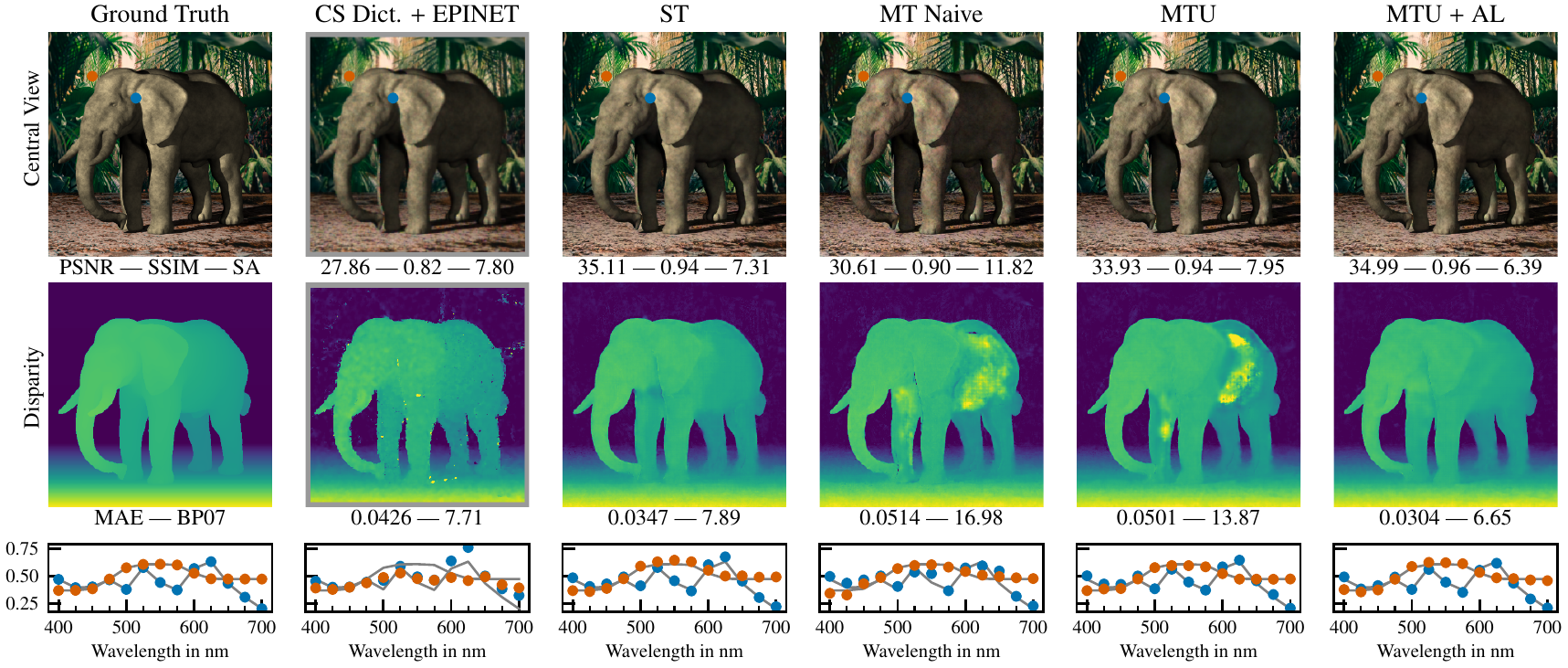}
	\caption{Performance comparison for full-sized prediction of the synthetic \textit{Elephant} challenge.
	In the multi-task case we consider MT with uncertainty (MTU)~\cite{Kendall2018:MultiTaskUncertainty} and the proposed MTU approach using an auxiliary loss (MTU + AL).
	Evaluation metrics PSNR in dB, SA in $^\circ$, MAE in px and BP07 in \%.
	The spectra of the two marked points are depicted in blue and orange together with the ground truth in grey.}
	\label{fig:eval-elephant}
	\vspace{-0.5mm}
\end{figure*}
\begin{figure*}
	\centering
	\includegraphics[scale=0.96]{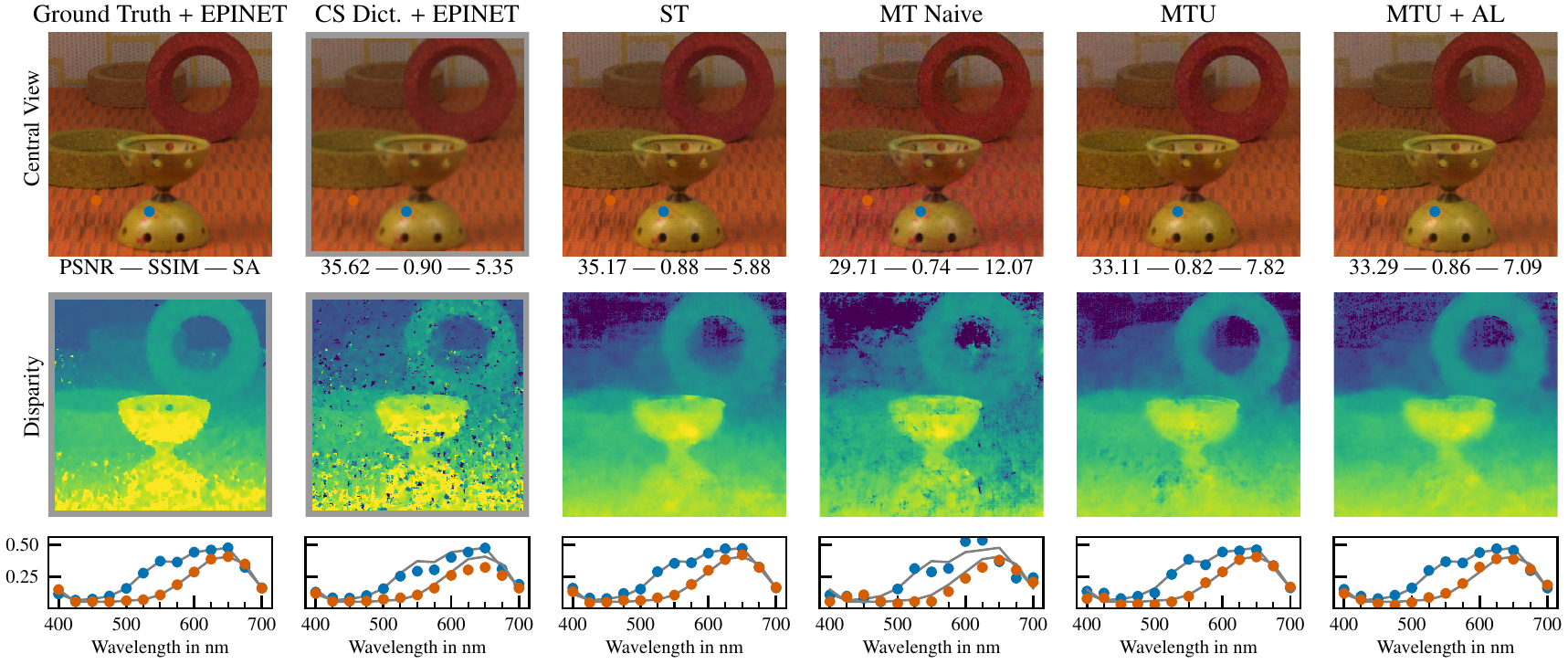}
	\caption{Performance comparison for full-sized prediction of the real-world \emph{Diavolo} scene. Captions are identical to Figure~\ref{fig:eval-elephant}. As no disparity ground truth is available, the prediction from the \emph{uncoded}, RGB-converted light field using EPINET~\cite{Shin2018:EPINET} is shown for comparison.}
	\label{fig:predict-real-diavolo}
	\vspace{-0.5mm}
\end{figure*}

\section{Conclusion}
\label{sec:conclusion}
We proposed a new method to infer light field properties from coded measurements directly to which we refer as principal reconstruction.
We reconstructed the central view and its aligned disparity map from spatio-spectrally coded light fields as taken by a light field camera with a spectrally coded microlens array achieving high quality results.
Despite the challenging hardware realisation, the camera design offers the great advantage of being compact and single-shot.

\textbf{Limitations and future work:}
The proposed NormGradSim approach is computationally and memory intensive.
We have experimented with gradient subsampling, using both static as well as stochastic subsets of the network weights to calculate the gradients.
However, the performance gains of the full gradient-based method were not achieved.
It is an interesting future investigation to employ dimensionality reduction methods, such as binary random projections, to the NormGradSim approach.
Moreover, we are interested in a superresolution extension of the proposed architecture.
Finally, optimization of the coding mask could further improve the reconstruction and will be investigated in the future.

\textbf{Acknowledgment:} This work was financed by the Baden-W\"urttemberg Stiftung gGmbH.
We acknowledge support by the state of Baden-Württemberg through bwHPC.

%
%

\clearpage
\clearpage
\bibliographystyle{ieee_fullname}
\bibliography{lit}

\clearpage
\clearpage
\begin{appendix}

\section{Network and Training Details}
\label{sec:network-details}

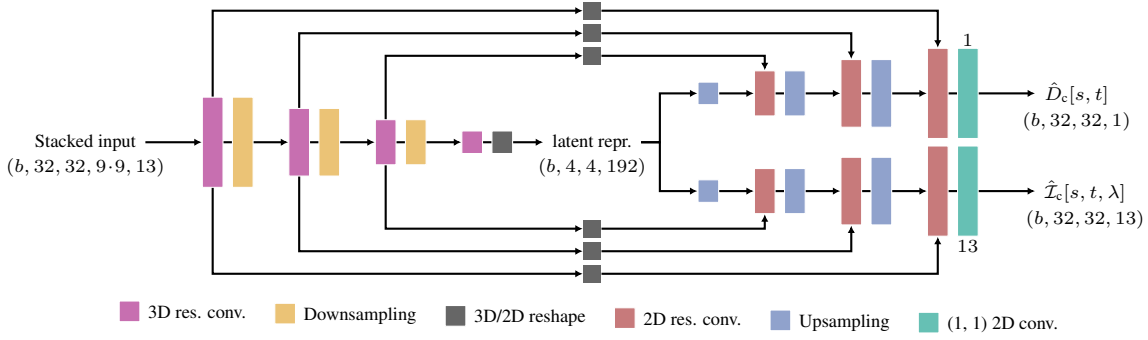
\begin{figure*}[t]
	\centering
	\begin{tikzpicture}[node distance = 0.75cm]
	\tikzstyle{every node}=[font=\scriptsize]
	
	\newcommand{\NETBLOCKHEIGHT}{1.2cm}
	\newcommand{\NETBLOCKWIDTH}{0.275cm}
	\newcommand{\NETBLOCKSEP}{0.35cm}
	
	\tikzstyle{netblockmaster} = [draw, minimum width=\NETBLOCKWIDTH, minimum height=\NETBLOCKHEIGHT]
	\tikzstyle{netblockslave} = [netblockmaster, xshift=-\NETBLOCKSEP]
	\tikzstyle{legend} = [draw, minimum width=\NETBLOCKWIDTH, minimum height=\NETBLOCKWIDTH, xshift=-5mm]
	
	\tikzstyle{sizefull} = [minimum height=\NETBLOCKHEIGHT]
	\tikzstyle{sizethreeq} = [minimum height=0.75*\NETBLOCKHEIGHT]
	\tikzstyle{sizehalf} = [minimum height=0.5*\NETBLOCKHEIGHT]
	\tikzstyle{sizeq} = [minimum height=0.25*\NETBLOCKHEIGHT]

	\tikzstyle{labelabove} = [yshift=1mm]
	\tikzstyle{labelbelow} = [yshift=-1mm]
	\tikzstyle{legendlabel} = [xshift=-6.25mm, yshift=-0.3mm]
	
	\tikzstyle{downsample} = [white, fill=kit-blue, fill opacity=0.6]
	\tikzstyle{residual3d} = [white, fill=kit-purple, fill opacity=0.6]
	\tikzstyle{downsample} = [white, fill=kit-orange, fill opacity=0.6]
	\tikzstyle{reshape} = [white, fill=black, fill opacity=0.6]
	\tikzstyle{upsample} = [white, fill=kit-blue, fill opacity=0.6]
	\tikzstyle{residual2d} = [white, fill=kit-red, fill opacity=0.6]
	\tikzstyle{final} = [white, fill=kit-green, fill opacity=0.6]

	\node (input) {Stacked input};
	\node[labelbelow] at (input.south) {$(b, 32, 32, 9\!\cdot\!9, 13)$};
	
	\node[right=of input, netblockmaster, residual3d, sizefull] (e1a) {};
	\node[right of=e1a, netblockslave, downsample, sizefull] (e1b) {};
	
	\node[right of=e1b, netblockmaster, residual3d, sizethreeq] (e2a) {};
	\node[right of=e2a, netblockslave, downsample, sizethreeq] (e2b) {};
	
	\node[right of=e2b, netblockmaster, residual3d, sizehalf] (e3a) {};
	\node[right of=e3a, netblockslave, downsample, sizehalf] (e3b) {};
	
	\node[right of=e3b, netblockmaster, residual3d, sizeq] (e4a) {};
	\node[right of=e4a, netblockslave, reshape, sizeq] (e4b) {};
	
	\node[right=of e4b, xshift=-3.5mm] (latent) {latent repr.};
	\node[labelbelow] at (latent.south) {$(b, 4, 4, 192)$};
	
	\node[right=of latent, yshift=0.65cm, netblockmaster, upsample, sizeq] (d11) {};
	
	\node[right of=d11, netblockmaster, residual2d, sizehalf] (d12a) {};
	\node[right of=d12a, netblockslave, upsample, sizehalf] (d12b) {};
	
	\node[right of=d12b, netblockmaster, residual2d, sizethreeq] (d13a) {};
	\node[right of=d13a, netblockslave, upsample, sizethreeq] (d13b) {};
	
	\node[right of=d13b, netblockmaster, residual2d, sizefull] (d14a) {};
	\node[right of=d14a, netblockslave, final, sizefull] (d14b) {};
	\node[labelabove] at (d14b.north) {$1$};
	
	\node[right=of d14b] (outdisp) {$\hat{D}_{\textrm{c}}[s, t]$};
	\node[labelbelow] at (outdisp.south) {$(b, 32, 32, 1)$};
	
	\node[right=of latent, yshift=-0.65cm, netblockmaster, upsample, sizeq] (d21) {};
	
	\node[right of=d21, netblockmaster, residual2d, sizehalf] (d22a) {};
	\node[right of=d22a, netblockslave, upsample, sizehalf] (d22b) {};
	
	\node[right of=d22b, netblockmaster, residual2d, sizethreeq] (d23a) {};
	\node[right of=d23a, netblockslave, upsample, sizethreeq] (d23b) {};
	
	\node[right of=d23b, netblockmaster, residual2d, sizefull] (d24a) {};
	\node[right of=d24a, netblockslave, final, sizefull] (d24b) {};
	\node[labelbelow, yshift=-0.25mm] at (d24b.south) {$13$};
	
	\node[right=of d24b] (outcentral) {$\hat{\mathcal{I}}_{\textrm{c}}[s, t, \lambda]$};
	\node[labelbelow] at (outcentral.south) {$(b, 32, 32, 13)$};
	
	\draw[-\myarrowhead, thick] (input)  -- (e1a);
	\draw[thick] (e1a)  -- (e1b);
	\draw[-\myarrowhead, thick] (e1b)  -- (e2a);
	\draw[thick] (e2a)  -- (e2b);
	\draw[-\myarrowhead, thick] (e2b)  -- (e3a);
	\draw[thick] (e3a)  -- (e3b);
	\draw[-\myarrowhead, thick] (e3b)  -- (e4a);
	\draw[thick] (e4a)  -- (e4b);
	\draw[-\myarrowhead, thick] (e4b)  -- (latent);
	
	\draw[-\myarrowhead, thick] (latent.east) -- +(0.25cm, 0) |- (d11);
	\draw[-\myarrowhead, thick] (d11)  -- (d12a);
	\draw[thick] (d12a)  -- (d12b);
	\draw[-\myarrowhead, thick] (d12b)  -- (d13a);
	\draw[thick] (d13a)  -- (d13b);
	\draw[-\myarrowhead, thick] (d13b)  -- (d14a);
	\draw[thick] (d14a)  -- (d14b);
	\draw[-\myarrowhead, thick] (d14b)  -- (outdisp);
	\draw[-\myarrowhead, thick] (latent.east) -- +(0.25cm, 0) |- (d21);
	\draw[-\myarrowhead, thick] (d21)  -- (d22a);
	\draw[thick] (d22a)  -- (d22b);
	\draw[-\myarrowhead, thick] (d22b)  -- (d23a);
	\draw[thick] (d23a)  -- (d23b);
	\draw[-\myarrowhead, thick] (d23b)  -- (d24a);
	\draw[thick] (d24a)  -- (d24b);
	\draw[-\myarrowhead, thick] (d24b)  -- (outcentral);
	
	\node[above of=latent, reshape, yshift=10mm] (r11) {};
	\draw[-\myarrowhead, thick] (e1a)  |- (r11);
	\draw[-\myarrowhead, thick] (r11) -| (d14a);
	
	\node[above of=latent, reshape, yshift=7mm] (r12) {};
	\draw[-\myarrowhead, thick] (e2a)  |- (r12);
	\draw[-\myarrowhead, thick] (r12) -| (d13a);
	
	\node[above of=latent, reshape, yshift=4mm] (r13) {};
	\draw[-\myarrowhead, thick] (e3a)  |- (r13);
	\draw[-\myarrowhead, thick] (r13) -| (d12a);
	
	\node[below of=latent, reshape, yshift=-10mm] (r21) {};
	\draw[-\myarrowhead, thick] (e1a)  |- (r21);
	\draw[-\myarrowhead, thick] (r21) -| (d24a);
	
	\node[below of=latent, reshape, yshift=-7mm] (r22) {};
	\draw[-\myarrowhead, thick] (e2a)  |- (r22);
	\draw[-\myarrowhead, thick] (r22) -| (d23a);
	
	\node[below of=latent, reshape, yshift=-4mm] (r23) {};
	\draw[-\myarrowhead, thick] (e3a)  |- (r23);
	\draw[-\myarrowhead, thick] (r23) -| (d22a);
	
	\node[below of=latent, xshift=-5.65cm, yshift=-1.5cm, legend, residual3d] (l1) {};
	\node[right of=l1, anchor=west, legendlabel] (l1label) {3D res. conv.\vphantom{p}};
	
	\node[right = of l1label, legend, downsample] (l2) {};
	\node[right of=l2, anchor=west, legendlabel] (l2label) {Downsampling\vphantom{p}};
	
	\node[right = of l2label, legend, reshape] (l3) {};
	\node[right of=l3, anchor=west, legendlabel] (l3label) {3D/2D reshape\vphantom{p}};
	
	\node[right = of l3label, legend, residual2d] (l4) {};
	\node[right of=l4, anchor=west, legendlabel] (l4label) {2D res. conv.\vphantom{p}};
	
	\node[right = of l4label, legend, upsample] (l5) {};
	\node[right of=l5, anchor=west, legendlabel] (l5label) {Upsampling\vphantom{p}};
	
	\node[right = of l5label, legend, final] (l6) {};
	\node[right of=l6, anchor=west, legendlabel] (l6label) {(1, 1) 2D conv.\vphantom{p}};

\end{tikzpicture}
	\caption{Schematic overview of the used dual-task encoder-decoder network.
	The shapes depicted correspond to the training data shapes.}
	\label{fig:network-large}
\end{figure*}

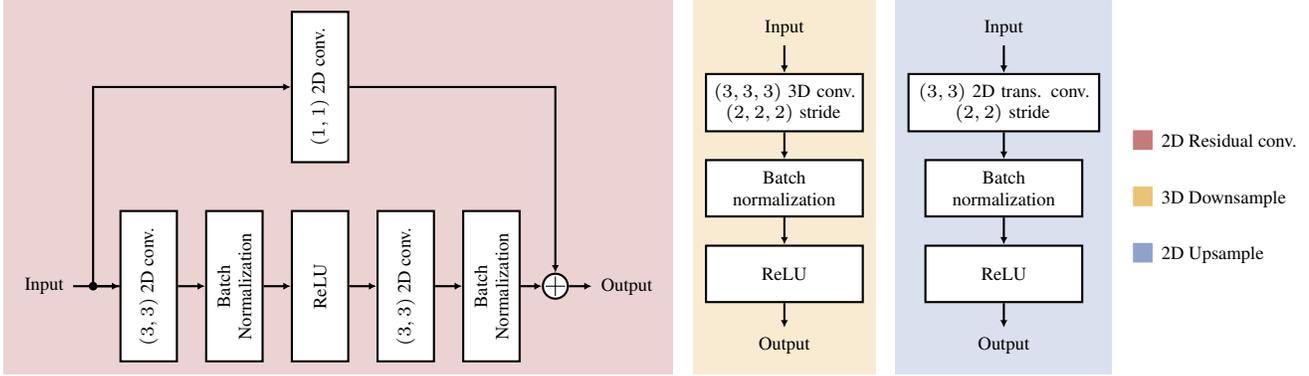
\begin{figure*}[t]
	\vskip 0pt
		\usetikzlibrary{backgrounds}
\begin{tikzpicture}[node distance = 0.75cm, background rectangle/.style={fill=kit-red!20}, show background rectangle]

\tikzstyle{every node}=[font=\scriptsize]
\tikzstyle{myFlow} = [rectangle, fill=white, anchor=west, minimum height=2cm, minimum width=0.75cm, draw=black,thick]
\tikzstyle{myAdd} = [draw =black , fill=white, circle , minimum size=1pt,thick]

\node (in) {Input};
\node[right of=in, xshift=-1mm, circle,fill=black, inner sep=0.78pt,outer sep=0pt, draw=black, minimum width=0.1cm, minimum height=0.1cm] (p0) {};
\node[right of = p0, myFlow, xshift=-4mm] (p1) {};
\node at (p1) {\rotatebox{90}{$(3, 3)$ 2D conv.}};
\node[right of = p1, myFlow] (p2) {};
\node at (p2) {\rotatebox{90}{\parbox{1.6cm}{\centering Batch\\Normalization}}};
\node[right of = p2, myFlow] (p3) {};
\node at (p3) {\rotatebox{90}{ReLU}};
\node[right of = p3, myFlow] (p4) {};
\node at (p4) {\rotatebox{90}{$(3, 3)$ 2D conv.}};
\node[right of = p4, myFlow] (p5) {};
\node at (p5) {\rotatebox{90}{\parbox{1.6cm}{\centering Batch\\Normalization}}};
\node[right of = p5, myAdd, xshift=1mm] (p6) {};
\draw[shorten <=0.5mm, shorten >=0.5mm] (p6.north) -- (p6.south);
\draw[shorten <=0.5mm, shorten >=0.5mm] (p6.east) -- (p6.west);
\node[right of = p6, xshift=2mm] (p7) {Output};

\node[above of=p3, myFlow, anchor=center, yshift=1.9cm] (res) {};
\node at (res) {\rotatebox{90}{$(1, 1)$ 2D conv.}};

\draw[thick] (in) -- (p1);
\draw[-\myarrowhead, thick] (p0) -- (p1);
\draw[-\myarrowhead, thick] (p1) -- (p2);
\draw[-\myarrowhead, thick] (p2) -- (p3);
\draw[-\myarrowhead, thick] (p3) -- (p4);
\draw[-\myarrowhead, thick] (p4) -- (p5);
\draw[-\myarrowhead, thick] (p5) -- (p6);
\draw[-\myarrowhead, thick] (p6) -- (p7);
\draw[-\myarrowhead, thick] (p0) |- (res);
\draw[-\myarrowhead, thick] (res) -| (p6);

\end{tikzpicture}
		\usetikzlibrary{backgrounds}
\begin{tikzpicture}[node distance = 0.75cm, background rectangle/.style={fill=kit-orange!20}, show background rectangle]

\tikzstyle{every node}=[font=\scriptsize]
\tikzstyle{myFlow} = [rectangle, fill=white, anchor=north, minimum width=1cm, minimum height=0.75cm, text centered, draw=black,thick,text width = 1.85cm]
    
\node (p1) {Input};
\node[below of = p1, myFlow, yshift=1.5mm] (p2) {$(3, 3, 3)$ 3D conv. $(2, 2, 2)$ stride};
\node[below of = p2, myFlow] (p3) {Batch\\ normalization};
\node[below of = p3, myFlow] (p4) {ReLU};
\node[below of = p4, yshift=-2mm] (p5) {Output};

\draw[-\myarrowhead, thick] (p1) -- (p2);
\draw[-\myarrowhead, thick] (p2) -- (p3);
\draw[-\myarrowhead, thick] (p3) -- (p4);
\draw[-\myarrowhead, thick] (p4) -- (p5);
\end{tikzpicture}
		\usetikzlibrary{backgrounds}
\begin{tikzpicture}[node distance = 0.75cm, background rectangle/.style={fill=kit-blue!20}, show background rectangle]

\tikzstyle{every node}=[font=\scriptsize]
\tikzstyle{myFlow} = [rectangle, fill=white, anchor=north, minimum width=1.85cm, minimum height=0.75cm, text centered, draw=black,thick,text width = 1.85cm]
    
\node (p1) {Input};
\node[below of = p1, myFlow, text width=2.3cm, yshift=1.5mm] (p2) {$(3, 3)$ 2D trans. conv.\\ $(2, 2)$ stride};
\node[below of = p2, myFlow] (p3) {Batch\\ normalization};
\node[below of = p3, myFlow] (p4) {ReLU};
\node[below of = p4, yshift=-2mm] (p5) {Output};

\draw[-\myarrowhead, thick] (p1) -- (p2);
\draw[-\myarrowhead, thick] (p2) -- (p3);
\draw[-\myarrowhead, thick] (p3) -- (p4);
\draw[-\myarrowhead, thick] (p4) -- (p5);

\end{tikzpicture}
		\usetikzlibrary{backgrounds}
\begin{tikzpicture}[node distance = 0.75cm]
\tikzstyle{every node}=[font=\scriptsize]
\newcommand{\TMPLEGENDWIDTH}{0.275cm}
\tikzstyle{upsample} = [draw, white, fill=kit-blue, fill opacity=0.6, minimum width=\TMPLEGENDWIDTH, minimum height=\TMPLEGENDWIDTH]
\tikzstyle{residual} = [draw, white, fill=kit-red, fill opacity=0.6, minimum width=\TMPLEGENDWIDTH, minimum height=\TMPLEGENDWIDTH]
\tikzstyle{downsample} = [draw, white, fill=kit-orange, fill opacity=0.6, minimum width=\TMPLEGENDWIDTH, minimum height=\TMPLEGENDWIDTH]
\tikzstyle{legendlabel} = [xshift=-6.25mm, yshift=-0.3mm]

\node (origin) {};
\node[residual, yshift=3cm] (res) at (origin) {};
\node[right of=res, anchor=west, legendlabel] {2D Residual conv.\vphantom{p}};
\node[below of=res, downsample] (down) {};
\node[right of=down, anchor=west, legendlabel] {3D Downsample};
\node[below of=down, upsample] (up) {};
\node[right of=up, anchor=west, legendlabel] {2D Upsample};

\end{tikzpicture}
	\vskip 1mm
	\caption{Schematic drawings of the used convolutional blocks. In the case of 3D residual convolution, all 2D convolutional blocks are replaced with 3D blocks with a kernel size of $(3,3,3)$ and $(1, 1, 1)$, respectively.}
	\label{fig:network-blocks}
\end{figure*}

The proposed network architecture, as shown in Figure~\ref{fig:network-large}, is built using the blocks shown in Figure~\ref{fig:network-blocks}.
For the encoder, the number of used filters per layer is doubled after every downsampling layer starting from 24 in the first residual block to 192 in the latent-space residual block.
All convolutions of the encoder are 3D convolutions with a kernel size of $(3, 3, 3)$, except for the residual convolutions with a kernel size of $(1, 1, 1)$.
Downsampling is performed using strided convolution with a $(2, 2, 2)$ stride.
The decoder is built symmetrically for both decoding paths, however 2D convolutions are used instead.
For upsampling, 2D transposed convolutions with a $(2, 2)$ stride are utilized.
In the last layer of each stream, a $(1, 1)$ convolution with either 13 or 1 features is performed to obtain the final shapes $(S, T, 13)$ and $(S, T, 1)$ for the reconstructed central view and disparity, respectively.
The encoder features are joined with the decoder features via the skip connections by concatenation.
In principle, it is straightforward to extend the encoder by an additional upsampling block to achieve superresolution in either one of the separate decoder paths, or to add (or replace) decoder paths to estimate different light field intrinsics.

In the case of the naive multi-task approach, we weigh each task equally, \ie we choose $w_i = 0.5,\,i=1,2$.
The single-task networks are obtained by setting $w_i = \delta_{ij}$.
An overview of the used loss weights is given in Table~\ref{tab:training-weights}.

\begin{table}
	\centering
	\caption{Comparison of the used multi-task weights $w_{\textrm{cv}}$, $w_{\textrm{disp}}$, and the auxiliary loss weights $w_{\textrm{aux}, j}$ for the investigated Methods.}
	\label{tab:training-weights}
	\vspace{1mm}
	\begin{tabular}{lrrr}
		\toprule
		\thead{Method} & \thead{$w_{\textrm{cv}}$} & \thead{$w_{\textrm{disp}}$} & \thead{$w_{\textrm{aux}, j}$}  \\
		\midrule
		ST Central View 	& \num{1.0} & \num{0.0} & -- \\
		ST Disparity 		& \num{0.0} & \num{1.0} & -- \\
		MT Naive 			& \num{0.5} & \num{0.5} & -- \\
		MT Uncertrainty		& adapt. & adapt. & -- \\
		MT GradNorm			& adapt. & adapt. & -- \\
		AL Gradient Similarity & \num{0.5} & \num{0.5} & adapt. \\
		AL NormGradSim & \num{0.5} & \num{0.5} & adapt. \\
		MTU + AL NormGradSim & adapt. & adapt. & adapt. \\
		\midrule
	\end{tabular}
\end{table}

For the main central view and disparity loss, we use the Huber loss.
The single element Huber loss is defined as
\begin{equation}
	{H}_\delta (e_i) = \begin{cases}
		e_i^2\, ,  &e_i < \delta \\
		2\delta \cdot (e_i - \frac{1}{2}\delta) \, ,  &\textrm{else}
	\end{cases}
\end{equation}
where $e_i = \lvert \myvec{y}_i - \hat{\myvec{y}}_i \rvert$ denotes the absolute prediction error of the $i$-th element for the vectorized estimate $\hat{\myvec{y}}$ with respect to the ground truth $\myvec{y}$.
The overall loss is then calculated as the mean of the element-wise Huber losses with $\delta=1$.

For the auxiliary loss methods we use the SSIM-based
\begin{equation}
	L_{\mathrm{SSIM}}(\mytensor{I}_{\textrm{c}}, \hat{\mytensor{I}}_{\textrm{c}}) = \frac{1}{2} \Big( 1 - \mathrm{SSIM}(\mytensor{I}_{\textrm{c}}, \hat{\mytensor{I}}_{\textrm{c}}) \Big)
\end{equation}
to enhance the spatial reconstruction of the central view.
Here, the SSIM is calculated channel-wise and averaged.
To enhance the spectral reconstruction, we use
\begin{equation}
	L_{\mathrm{CS}}(\mytensor{I}_{\textrm{c}}, \hat{\mytensor{I}}_{\textrm{c}})
	= \frac{1}{2} \Bigg( 1 -
	\frac{\langle \mytensor{I}_{\textrm{c}}, \hat{\mytensor{I}}_{\textrm{c}}\rangle_{\lambda}}{\lVert\mytensor{I}_{\textrm{c}} \rVert_{\lambda}\!\cdot\! \lVert \hat{\mytensor{I}}_{\textrm{c}}\rVert_{\lambda}}
	\Bigg)
\end{equation}
which is based on the spectral cosine similarity.
The similarity is calculated along the spectral axis and spatially averaged.
For the disparity task, we employ a total variation-based smoothness enhancing auxiliary loss.
The used loss is similar to the one introduced by Repala~\etal~\cite{Repala2019:disparity-smoothness} but instead of weighing the local disparity gradients with the gradients of the central view, which is problematic in textured but constant-disparity regions, we calculate
\begin{equation}
	L_{\mathrm{TV}}(\myvec{D}_{\textrm{c}}, \hat{\myvec{D}}_{\textrm{c}})
	= \Big\lvert \partial_{x} \hat{\myvec{D}}_{\textrm{c}} \cdot\mathrm{e}^{-\lVert \partial_{x} \myvec{D}_{\textrm{c}}\rVert} \Big\rvert
	+ \Big\lvert \partial_{y} \hat{\myvec{D}}_{\textrm{c}} \cdot\mathrm{e}^{-\lVert \partial_{y} \myvec{D}_{\textrm{c}}\rVert} \Big\rvert
\end{equation}
using the disparity ground truth gradients.
As opposed to the standard total variation, the exponential weighting should reduce edge-related artifacts.
Furthermore, we use the disparity normal similarity loss~\cite{Hu2019:Revisiting-single-image-depth}
\begin{equation}
	L_{\mathrm{NS}}(\myvec{D}_{\textrm{c}}, \hat{\myvec{D}}_{\textrm{c}})
	= \frac{1}{2} \Bigg( 1 -
	\frac{\langle \myvec{n}, \hat{\myvec{n}}\rangle}{\lVert \myvec{n} \rVert \!\cdot\! \lVert \hat{\myvec{n}}\rVert}
	\Bigg)
\end{equation}
which is spatially averaged.
The normals are calculated as
\begin{equation}
	\myvec{n} = [-\partial_{x} \myvec{D}_{\textrm{c}}, -\partial_{y} \myvec{D}_{\textrm{c}}, 1]^{\mathrm{T}}\,.
\end{equation}
Finally, layers that are preceded by batch normalization are regularized during training using the $l_2$-norm of the weights (weight decay).

The networks were trained for 170\,epochs with a mini-batch size of 128 using the YOGI optimizer~\cite{Reddi2018:Yogi-optimizer} and a sigmoid learning rate decay from $5\times 10^{-3}$ to $1\times 10^{-4}$ which has shown to be superior to linear, exponential, and step decay in preliminary experiments.

All learning-based methods (the proposed approach, EPINET, and the dictionary learning) were trained using a single \SI{32}{GB} Nvidia Tesla V100 GPU and 10 cores with \SI{96}{GB} RAM of a shared GPU computing node.
Inference with the deep learning methods was performed with the same hardware.
The full-size inference of the synthetic dataset challenges as well as the real-world data in the case of the CS-based methods could not be carried out on the GPU due to the large memory requirements.
Instead, they were carried out using a 40-core computing node with \SI{192}{GB} of RAM.
Training of the proposed methods took between 4 and 7 days, depending on the used training strategy.
The training of EPINET took about 1\,day.
The dictionary training converged after 1 epoch in about 2 days.

\section{Compressed Sensing-based Reconstruction}
\label{sec:compressed-sensing}

There are several approaches to solve the general compressed sensing reconstruction problem
\begin{equation}\label{eq:compessred-sensing-general-rep}
\min_{\myvec{\alpha}}\,\lVert \myvec{\alpha} \rVert_0 \quad \textrm{subject to }  \lVert \myvec{l}^{*} - \myvec{M}\myvec{\Psi}\myvec{\alpha} \rVert_2 < \epsilon \,.
\end{equation}
Here, $\myvec{\Psi}\myvec{\alpha} = \myvec{l} \in \mathbb{R}^{U\cdot V\cdot S\cdot T\cdot \Lambda}$ and $\myvec{l}^{*} \in \mathbb{R}^{U\cdot V\cdot S\cdot T}$ denote the vectorized versions of $\mytensor{L}$ and $\mytensor{L}^{*}_{\textrm p}$, respectively, $\myvec{\Psi}$ denotes a basis (or frame) in which $\myvec{l}$ has the sparse representation $\myvec{\alpha}$ and $\myvec{M}$ corresponds to the measurement matrix.

In general, the minimization of the non-convex $l_0$-pseudo norm is NP-hard~\cite{Eldar2012:Compressed-sensing}.
While greedy methods, that directly tackle the $l_0$-norm minimization, exist, we focus on the so-called convex relaxation methods here.
When the restricted isometry property of the measurement matrix $\myvec{M}$ and the basis or frame $\myvec{\Psi}$ is fulfilled, the optimization problem~\eqref{eq:compessred-sensing-general-rep} is equivalent to the convex LASSO problem~\cite{Eldar2012:Compressed-sensing}
\begin{equation}\label{eq:compessred-sensing-relaxed}
\min_{\myvec{\alpha}}\,\lVert \myvec{l}^{*} - \myvec{M}\myvec{\Psi}\myvec{\alpha} \rVert_2^2 + \lambda \lVert \myvec{\alpha} \rVert_1
\end{equation}
for some coupling constant $\lambda$.
In the case of light fields, the compressed sensing reconstruction is challenging due to the large dimensionality.
This becomes even more severe in the case of spectral light fields.
For example, even for a small light field patch of shape $(9, 9, 32, 32, 13)$, which corresponds to the size of the used light fields in our test dataset (\cf~Section~4.1 of the main paper), the basis $\myvec{\Psi}$ alone requires more than \SI{4.6}{TB} of memory at \SI{32}{bit} resolution.
This is unfeasible even for large-scale computers, not even considering the case of full-sized light fields.
The same issue is generally also true for the measurement matrix $\myvec{M}$.
However, in our particular case, the coding takes a simple diagonal form $\myvec{M} = \diag(\myvec{m})$ when not performing the projection of the spectral dimension.
While this increases the memory requirements of the coded light field (because it is not spectrally projected) the memory requirements of the coding mask is reduced drastically.
Furthermore, the coding can now be expressed as a simple element-wise multiplication instead of a matrix multiplication reducing the computational effort:
\begin{equation}
	\myvec{M}\myvec{\Psi}\myvec{\alpha} = \myvec{m} \odot \myvec{\Psi}\myvec{\alpha}\,.
\end{equation}
Still, to overcome the challenges regarding the basis $\myvec{\Psi}$, we consider the following approaches.

\subsection{DCT-based Reconstruction}
Using the 5D-DCT as a basis for the spectral light field reconstruction and a functional optimization approach, explicitly saving the basis matrix $\myvec{\Psi}$ can be avoided.
The minimization problem~\eqref{eq:compessred-sensing-relaxed} can be rewritten as
\begin{equation}
	\min_{\myvec{\alpha}}  f(\myvec{\alpha}) + \lambda \lVert \myvec{\alpha} \rVert_1
\end{equation}
where $f(\myvec{\alpha}) = \lVert \myvec{l}^{*} - \myvec{m} \odot \myvec{\Psi}\myvec{\alpha} \rVert_2^2$ is convex and differentiable.
We can explicitly calculate the derivative
\begin{align}
	\nabla f(\myvec{\alpha})
	&= 2\big(
	\myvec{\Psi}^{\mathrm{T}} \myvec{M}^{\mathrm{T}}\myvec{M} \myvec{\Psi} \myvec{\alpha}
	- \myvec{\Psi}^{\mathrm{T}} \myvec{M} \myvec{l}^{*} \,\big)\notag\\
	&= 2\big(
	\myvec{\Psi}^{\mathrm{T}} \myvec{m} \odot  \myvec{\Psi} \myvec{\alpha}
	- \myvec{\Psi}^{\mathrm{T}} \myvec{m} \odot \myvec{l}^{*} \,\big)\,.
\end{align}
Here, we used that $\myvec{M}^{\mathrm{T}} = \myvec{M} = \diag(\myvec{m})$ and that $\myvec{M}$ is an orthogonal projection, \ie $\myvec{M}^2 = \myvec{M}$.
This makes the optimization suitable for the limited-memory BFGS optimization~\cite{Byrd:1995:LBFGS} in the orthant-wise limited-memory quasi-Newton variant~\cite{Andrew:2007:OWL-QN-LFBGS}.
Using a functional (non matrix-based) implementation of the 5D-DCT $\myvec{\Psi}$ and its inverse transform $\myvec{\Psi}^{\mathrm{T}}$, this approach becomes feasible even for full-sized light field reconstruction.
In our case, the full-sized reconstruction of spectral light fields with shape $(9, 9, 512, 512, 13)$ requires about \SI{96}{GB} of RAM.
In principle, other basis transforms, such as the discrete wavelet transform or the signal-adapted wavelet packet transform, are also possible.
However, due to the small angular resolution of the light fields, the DCT has been argued to perform comparable or better than the wavelet-based methods, especially for large compression ratios~\cite{Baechle2021:Wavelet-packets}.
Since the DCT is computationally efficient and straightforward to implement, we are choosing it here.

On the other hand, the DCT as well as wavelet-based transforms do not explicitly consider the light field geometry.
While some light field-specific bases have been discussed, this is still an open and ongoing research topic.
To make use of the light field geometry for a sparse representation, we consider the following dictionary learning approach.

\subsection{Dictionary-based Reconstruction}
Another possibility for compressed sensing-based reconstruction is to use a signal-adapted dictionary (or frame).
Here, the goal is to find a dictionary $\myvec{D} \in \mathbb{R}^{N\times kN}$, where $N = U\!\cdot\!V\!\cdot\!S\!\cdot\!T\!\cdot\!\Lambda$ is the vectorized light field's dimension and $k$ is the dictionary overcompleteness, such that the (possibly approximate) light field representation $\myvec{\alpha} \in \mathbb{R}^{kN}$ with
\begin{equation}
	\myvec{l} = \myvec{D}\myvec{\alpha}
\end{equation}
is sparse.
The columns $\myvec{d}_i, i = 1, \dots, kN$, of $\myvec{D}$ are called the atoms of the dictionary.
Obviously, obtaining $\myvec{\alpha}$ from a given light field~$\myvec{l}$ is an inverse problem and may only be solved approximately.
To learn a dictionary $\myvec{D}$ from a training dataset $\myvec{L} \in \mathbb{R}^{N \times B}$ containing $B$ spectral light fields, the joint optimization problem
\begin{align}\label{eq:dictionary-learning-general}
&\min_{\myvec{D}, \myvec{A}}\,  \lVert \myvec{L} - \myvec{D}\myvec{A} \rVert_{2}^2 + \lambda \lVert \myvec{A} \rVert_1
	\,,\notag\\
&~\textrm{s. t.}\quad \myvec{1} \odot {\myvec{D}^{\mathrm{T}}\myvec{D}} = \myvec{1}\,,
\end{align}
has to be solved.
Here, $\myvec{A} \in \mathbb{R}^{kN \times B}$ denotes the sparse representation of all light fields in the training dataset.
The constraint on $\myvec{D}$ ensures that all atoms are normalized.
We perform the optimization in an alternating manner, separating the non-convex problem~\eqref{eq:dictionary-learning-general} into two convex sub-problems for $\myvec{D}$ and $\myvec{A}$ separately~\cite{Mairal2009:online-dictionary-learning}.
Further, since the used training dataset contains over \SI{200}{GB} of light field data, we employ mini-batch Stochastic Gradient Descent (SGD) optimization.
Hence, for each optimization iteration step, first a sparse representation $\myvec{A}$ of the light field batch is estimated using a fixed dictionary $\myvec{D}$, which is initialized using a truncated normal distribution and succeeding atom normalization.
There are several approaches to estimate the sparse decomposition.
Here, we use the Fast Iterative Shrinkage/Thresholding Algorithm (FISTA)~\cite{Beck2009:FISTA} to solve
\begin{align}
&\min_{\myvec{A}}\,  \lVert \myvec{L} - \myvec{D}\myvec{A} \rVert_{2}^2 + \lambda \lVert \myvec{A} \rVert_1 \,.
\end{align}
In the second step, the sparse representation is fixed, and the dictionary optimization
\begin{align}
&\min_{\myvec{D}}\,  \lVert \myvec{L} - \myvec{D}\myvec{A} \rVert_{2}^2 \,,\notag\\
&~\textrm{s. t.}\quad \myvec{1} \odot {\myvec{D}^{\mathrm{T}}\myvec{D}} = \myvec{1}\,,
\end{align}
is performed updating the dictionary atoms using SGD and succeeding atom normalization.

Over a fixed basis approach, the dictionary has the advantage of implicitly taking the light field geometry and redundancy into account.
However, one still faces the problem of large dimensionality.
Due to the dictionary overcompleteness, the problem is even slightly more severe than for the basis decomposition.
Therefore, we train the dictionary with light field atoms of shape $(5, 5, 8, 8, 13)$.
To encode an input light field using the dictionary, the light field is first patched into the corresponding shape, using a spatial overlap of $(4, 4)$ and an angular overlap of $(1, 1)$.
To avoid edge defects, overlapping patches are averaged when de-patching.
We decided to not employ patching in the spectral domain as the characteristics of the different spectral bands are assumed to be quite different.
The shape of the light field atoms was chosen as large as possible while still resulting in a manageable dictionary size.
Using a dictionary overcompleteness of $k=2$, which in previous works has been argued to be suitable for light field dictionaries~\cite{Marwah2013}, the dictionary is roughly \SI{3.5}{GB} in size.
With the additional memory requirements of the gradient backpropagation and light field patching, this was just small enough to perform the dictionary learning on a \SI{32}{GB} Nvidia Tesla V100 GPU.
Note, that the atom shape $(5, 5, 8, 8, 13)$, while seeming small in the context of light fields, is actually comparably high dimensional.
For image-based dictionary learning, this corresponds to atoms of shape $(144, 144)$, since $144\!\cdot\!144 \approx 5\!\cdot\!5\!\cdot\!8\!\cdot\!8\!\cdot\!13$.
However, even recent methods specifically adapted for high dimensional image dictionary learning only reach a feasible atom size of $(64, 64)$~\cite{Sulam2016:Trainlets}.

\section{Spectral Light Field Camera Calibration}
\label{app:camera-calibration}
\begin{figure}
	\centering
	\includegraphics[height=3cm]{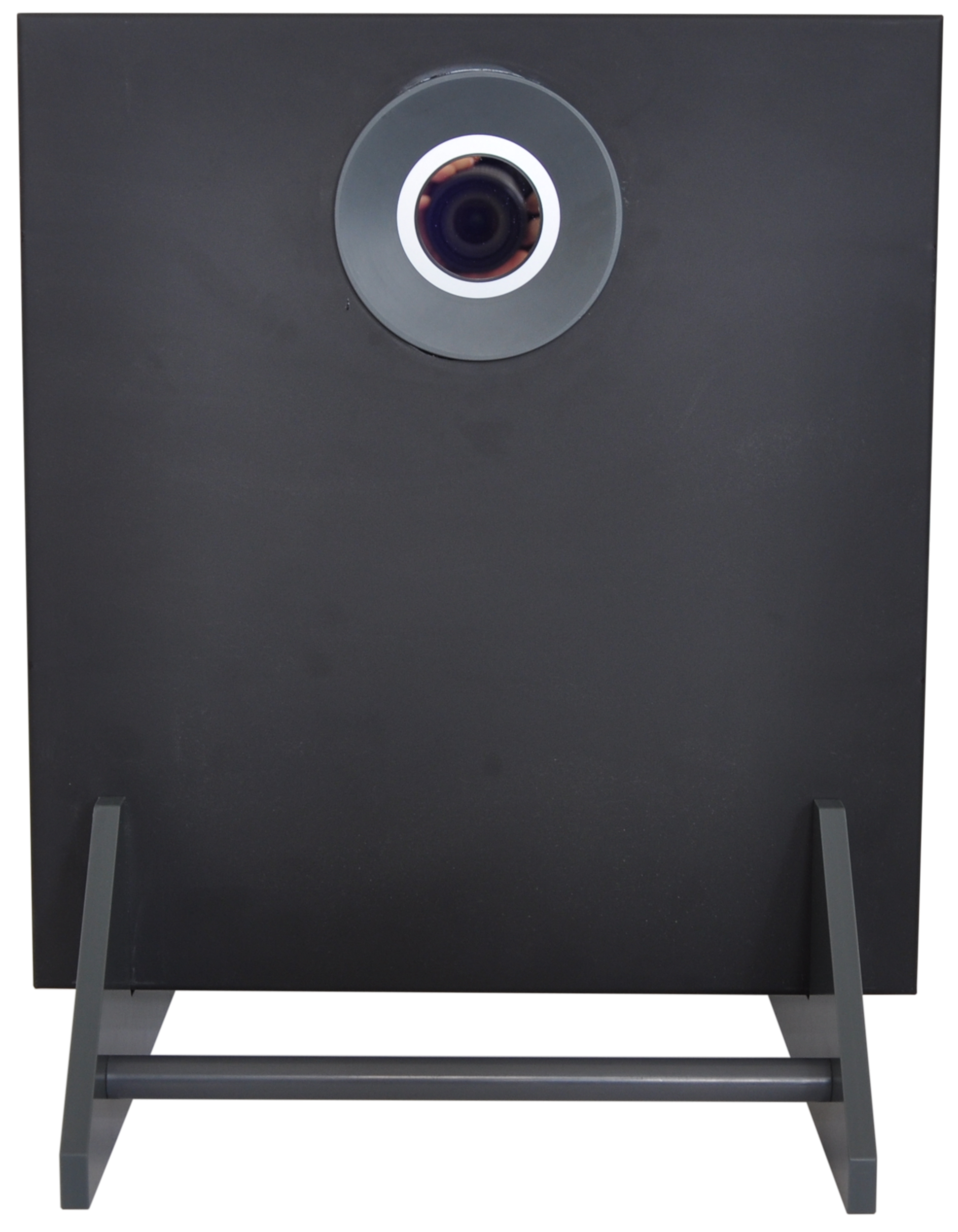}
	\hspace{3mm}
	\includegraphics[height=3cm]{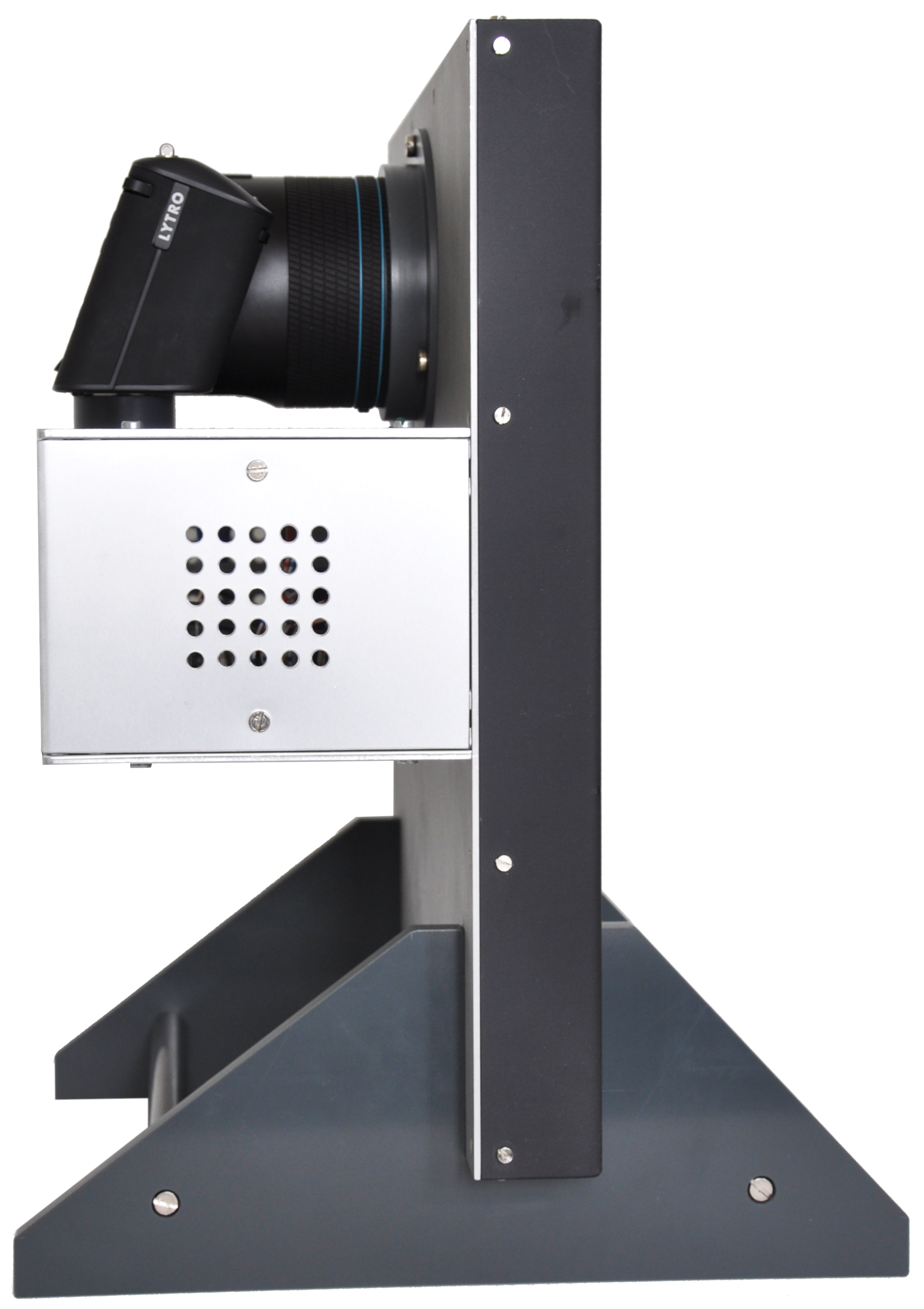}
	\includegraphics[height=3cm]{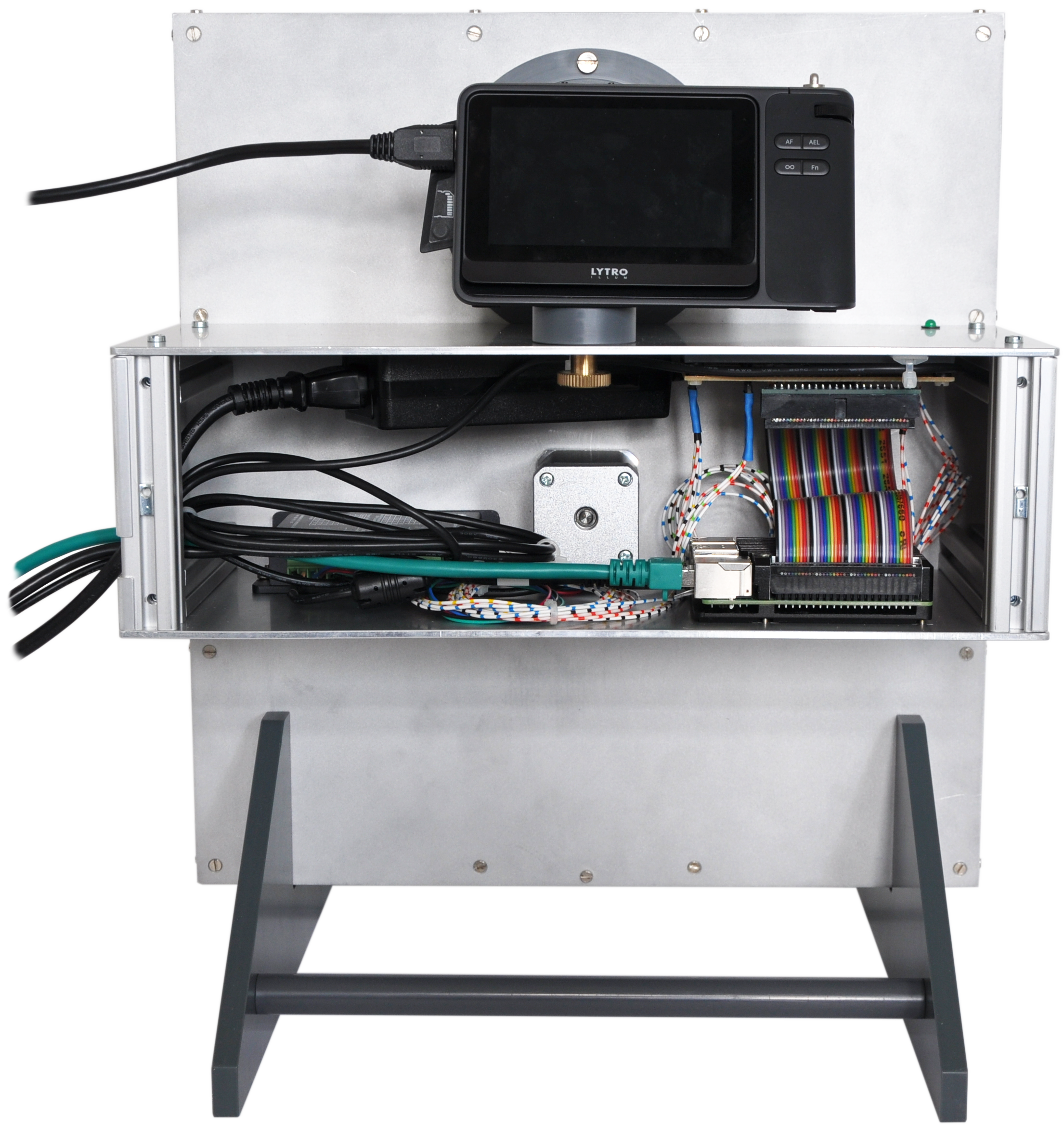}
	\vspace{2mm}
	\caption{Our custom spectral light field camera, using a Lytro Illum camera and a filter wheel, controlled by a Raspberry Pi.}
	\label{fig:camera-prototype}
\end{figure}

Our spectral light field camera is composed of a Lytro Illum camera and a custom-built housing holding a filter wheel with 13 spectral bandpass filters.
The filter wheel is flange-mounted onto a stepper motor which is controlled, together with the camera, by a Raspberry Pi.
A picture of our camera is shown in Figure~\ref{fig:camera-prototype}.

The spectral light field $\mathcal{L}[u, v, s, t, \lambda]$ is measured in a spectrally scanning fashion where for each spectral channel a light field is captured by the Lytro Illum camera employing the corresponding spectral bandpass filter.
Usually, the individual channels of a spectrally scanning camera are obtained using identical exposure times.
However, this results in a strongly channel-dependent signal-to-noise ratio (SNR) due to the different spectral sensitivities which are determined by the filter characteristics, the quantum efficiency of the sensor, as well as the used light source.
In our case, this effect is quite severe since we are employing regular daylight photo studio illuminants.
To overcome this problem, the spectral channels are captured with their individually optimal exposure time.
For example, in our case this results in very short exposures in the green channels and very long exposures of up to \SI{2}{\second} in the UV and NIR channels.
This approach has the drawback that the camera needs to be radiometrically calibrated which is more cumbersome than the regular white balancing and de-vignetting.
However, the radiometric calibration has the advantage that the additional Bayer-pattern RGB filters that are present in the Lytro Illum camera as well as the vignetting of the main lens (which is non-detachable) as well as the microlenses can be calibrated at once.

First, the camera's dark signal properties are estimated.
Following the EMVA 1288 standard, the mean value of the Poisson-distributed dark signal $\mathtt{d}$ is given by
\begin{align}\label{eq:dark-signal}
\mu_{\mathtt{d}} &= \mu_{\mathtt{d}, 0} +  \mu_{\mathtt{I}}\,t \,.
\end{align}
The mean dark signal $\mu_{\mathtt{d}}$ is linear in the exposure time $t$ with offset $\mu_{\mathtt{d}, 0}$ and slope $\mu_{\mathtt{I}}$ which is called the (mean) dark current.
The offset and slope can be estimated by measuring an exposure series of dark images, \ie images without any illumination and varying exposure time.
The mean values $\mu_{\mathtt{d}}$	are approximated using the sample mean with a sample size of 10, \ie the exposure series is repeatedly measured and averaged.
The mean offset and dark current are estimated via a simple linear least-squares regression of the averaged exposure series using~\eqref{eq:dark-signal}.
Note that this calibration has to be performed individually for every used camera gain.
For this reason, we fix the camera gain to ISO~80 for all measurements.

\begin{figure}[t]
	\centering
	\includegraphics{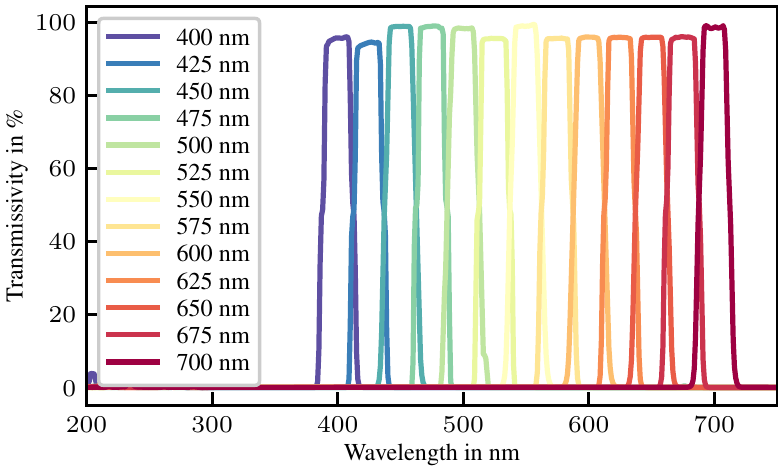}
	\caption{Transmissivity of the 13 used spectral bandpass filters.}
	\label{fig:filter_transmissivity}
\end{figure}

To radiometrically calibrate the camera, an exposure series of spectral bright images is collected.
Here, a bright image refers to an image of the light source taken through an optical diffuser, achieving an almost ideal diffuse white scene.
Following the EMVA 1288 standard, each pixel should obey the linear camera model:
\begin{align}
	\mu_{\mathtt{g}}
	&= \mu_{\mathtt{d}} + K \eta \frac{\lambda A}{h c} E t \notag\\
	&= \mu_{\mathtt{d}, 0} + (\alpha + \mu_{\mathtt{I}})\,t \,. \label{eq:emva-linear-model}
\end{align}
Here, $\mathtt{g}$ corresponds to the stochastic pixel grey value.
The quantum efficiency is denoted by $\eta$ and the system gain by $K$.
The term $\frac{\lambda A}{h c} E$ corresponds to the mean number of photons reaching the pixel (depending on the light source and the used lenses) and is unified in the scalar $\alpha$ which incorporates all spectral characteristics of the light source.
Overall, the mean pixel grey value is linear in the exposure time.
Now, in our case the camera employs additional spectral filters as well as a Bayer-pattern RGB sensor.
Since the filters have the same effect on the incoming photons as the sensor quantum efficiency, \ie, a photon is either transmitted with a certain probability or not, the filters can be multiplicatively incorporated into~\eqref{eq:emva-linear-model} via
\begin{align}
	\mu_{\mathtt{g}}
	&= \mu_{\mathtt{d}, 0} + (b_{n}\,\varphi_{k}\,\alpha + \mu_{\mathtt{I}})\,t \,,\\
	n &= 0, 1, 2 \,,&\text{(RGB filter)} \notag  \\
	k &= 0, 1, \dots 12\,,&\text{(spectral filter)} \notag
\end{align}
where $b_n$ denotes the effective RGB filter and $\varphi_{k}$ the spectral filter transmissivity.
Therefore, we can write the linear model for an abitrary spectral pixel $(i,j,k)$ in a tensor-like fashion:
\begin{align}
	\mu_{\mathtt{g},ijkl}
	&= \mu_{\mathtt{d}, 0} + (b_{n}\,\varphi_{k}\,\alpha_{ij} + \mu_{\mathtt{I}})\,t_{l} \,,\quad\textrm{or}\\
	\tilde{\mu}_{\mathtt{g},ijkl}
	&\coloneqq \mu_{\mathtt{g},ijkl} - \mu_{\mathtt{d}, 0} - \mu_{\mathtt{I}}\,t_{l}
	= b_{n(i,j)}\,\varphi_{k}\,\alpha_{ij}t_{l} \,. \label{eq:responsivity-signal-model}
\end{align}
The index $n(i,j)$ of the used RGB filter is uniquely determined by the pixel position $(i,j)$.
We now factorize the linear dependence into its spatial and spectral components, \ie we write
\begin{align}
	\tilde{\mu}_{\mathtt{g},ijkl} = v_{ij}r_k^{(n)}\,t_{l} \,.
\end{align}
Here, $v_{ij}$ denotes all spatial dependencies such as the natural and mechanical main lens and microlens vignetting, and $r_{k}$ denotes the spectral responsivity which depends on the pixel's filter type $n \in \{\mathrm{R}, \mathrm{G}, \mathrm{B}\}$ and the used bandpass filter $k$.
The goal is now to estimate $v_{ij}$ and $r_{k}^{(n)}$ from the measured spectral bright image exposures series.

\begin{figure*}[t]
	\begin{minipage}[t]{0.5\linewidth}
		\centering \vskip 0pt
		\includegraphics{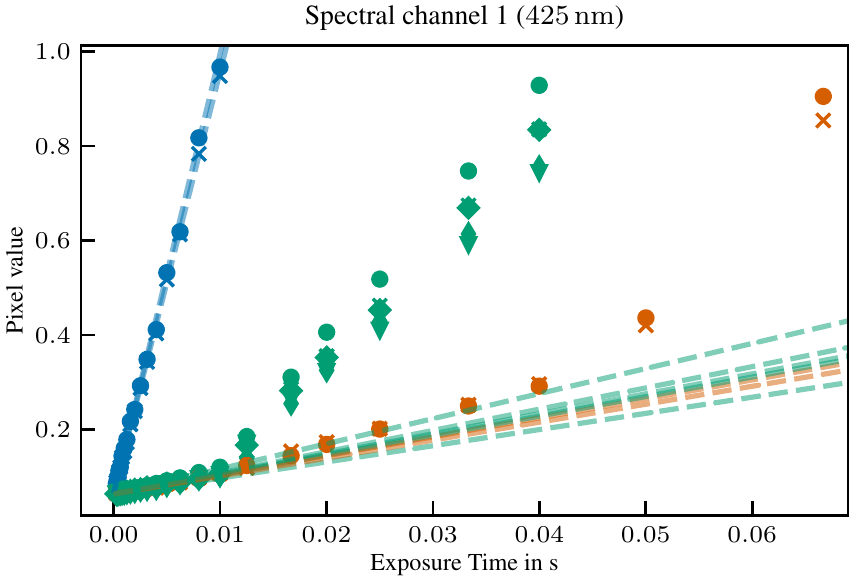}
	\end{minipage}
	\begin{minipage}[t]{0.5\linewidth}
		\centering \vskip 0pt
		\includegraphics{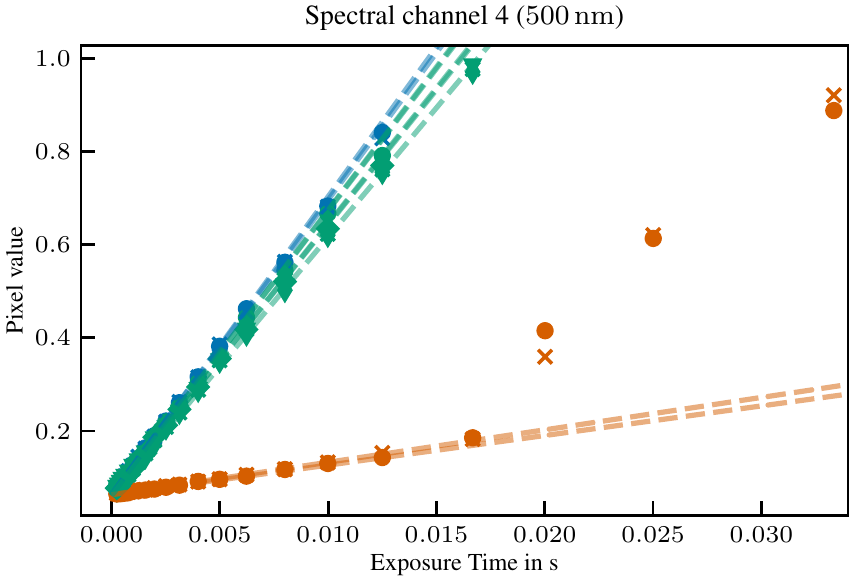}
	\end{minipage}\\[2mm]
	\begin{minipage}[t]{0.5\linewidth}
		\centering \vskip 0pt
		\includegraphics{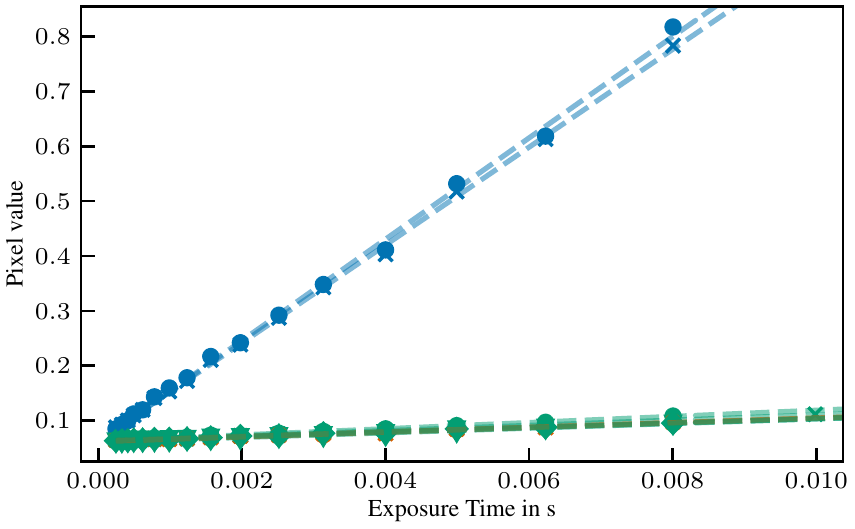}
	\end{minipage}
	\begin{minipage}[t]{0.5\linewidth}
		\centering \vskip 0pt
		\includegraphics{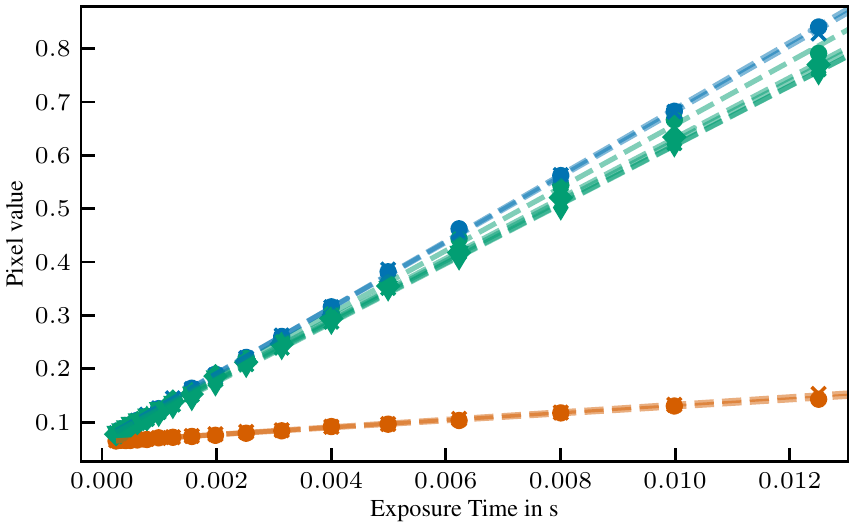}
	\end{minipage}\vspace{2mm}
	\caption{A $9\times9$ pixel crop from the spectral bright image exposure series.
	The individual pixel measurements are colored corresponding to their respective RGB filter type.
	Oversaturated pixels are not shown.
	Top row: unfiltered exposure series with linear fit of the first 10 exposure measurements.
	Bottom row: filtered with neighbor overexposure compensation and linear fit of the full data.}
	\label{fig:exposure-series}
\end{figure*}
\begin{figure*}
	\begin{minipage}[t]{0.25\linewidth}
		\centering \vskip 0pt
		\includegraphics[width=0.9\linewidth, interpolate=false]{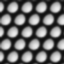}
	\end{minipage}
	\begin{minipage}[t]{0.25\linewidth}
		\centering \vskip 0pt
		\includegraphics[width=0.9\linewidth, interpolate=false]{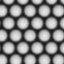}
	\end{minipage}
	\begin{minipage}[t]{0.5\linewidth}
		\centering \vskip 0pt
		\includegraphics{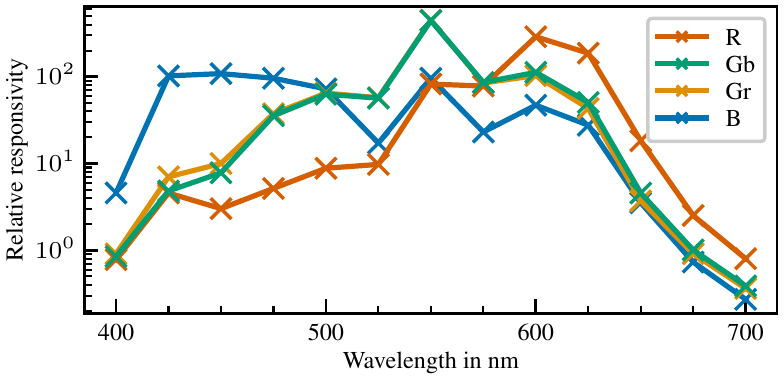}
	\end{minipage}
	\vspace{1mm}
	\caption{Estimated vignetting (top left and center crop from full image) and spectral responsivity.}
	\label{fig:exposure-series-fit}
\end{figure*}

In order to process the full spectral exposure series tensor $\tilde{\mu}_{\mathtt{g},ijkl}$ at once, which makes GPU acceleration (\eg via PyTorch or TensorFlow) straight-forward, certain care has to be taken to mask out overexposed pixels.
Since the exposure series is measured for all spectral channels $k$, overexposure is unavoidable.
For example, the green channels are much more sensitive than the NIR or UV channels and will therefore saturate much more quickly.
However, overexposure is easy to handle by masking out all pixels with pixel value larger than $\num{0.985}$, which corresponds to the threshold given by the four least significant bits of the \SI{10}{bit} sensor\footnote{$(1111110000)_2/(1111111111)_2 = 1008/1023 \approx \num{0.985}$.}.
However, this simple masking is not sufficient.
As is shown in Figure~\ref{fig:exposure-series}, overexposed pixels influence neighboring pixels.
In CCD sensors, the charges from saturated pixels overflow to neighboring pixels, in particular along the line at which the CCD sensor is read out.
In the top left of Figure~\ref{fig:exposure-series} we observe that the blue pixels saturate first, leading to a change of slope of the green pixels' sensitivity, since they are now also registering the overflown charges from the blue pixels.
This is referred to as \emph{blooming} in CCD sensors.
However, the red pixels' sensitivity only changes slightly.
This is likely because the red and blue pixels are only neighbored diagonally, whereas the green pixels are direct neighbors to both blue and red pixels.
So when also the green pixels saturate, the red pixels' sensitivity also changes abruptly.
To obtain a reliable estimate of the true sensitivity, it is clear that these measurements have to be masked out.
We do this by using the available saturation mask and extend it to direct neighbors as well as neighbors in the \SI{5}{px} neighborhood along the line along which the CCD sensor is read out.
The filtered results are shown in Figure~\ref{fig:exposure-series} (bottom).
Furthermore, since the exposure times of the camera are logarithmically more densely sampled at short exposure times, we weigh the measurements logarithmically.
Finally, we obtain an estimate of the vignetting and responsivity using the model~\eqref{eq:responsivity-signal-model} and a weighted least-square fit.
Since the model parameters are coupled, an analytic solution is not available.
We minimize the squared error using PyTorch and the Adam optimizer.
An example result of the fitted vignetting and responsivity is given in Figure~\ref{fig:exposure-series-fit}.
Since the full spectral exposure series cannot be loaded onto the GPU at once (the measurements add up to about \SI{65}{GB} per camera configuration), the measurements are spatially patched using 64 windows with \SI{50}{\percent} overlap.

The individually captured (monochromatic) raw light fields have then to be decoded, and concatenated to obtain the spectral light field.
To decode the raw sensor images to a light field, the camera has to be geometrically calibrated.
That is, the regular MLA grid has to be estimated from the individual ML centers.
Here, we use the calibration and decoding scheme proposed by Schambach and Puente Le\'on~\cite{Schambach2020} which itself is a vignetting-aware refinement of the well established decoding method by Dansereau \etal~\cite{Dansereau2013}.
For each focus and zoom setting, a regular grid is estimated to approximate the individual ML centers.
The raw image is then aligned with the regular grid and patched into a light field.
After patching, the obtained light field has to be resampled from a hexagonal to a rectangular grid.
Since this is well established it is not elaborated in more detail here and the interested reader is referred to the original literature.

	\section{Additional Results}
\label{sec:addtional-results}
\begin{figure*}
    \includegraphics{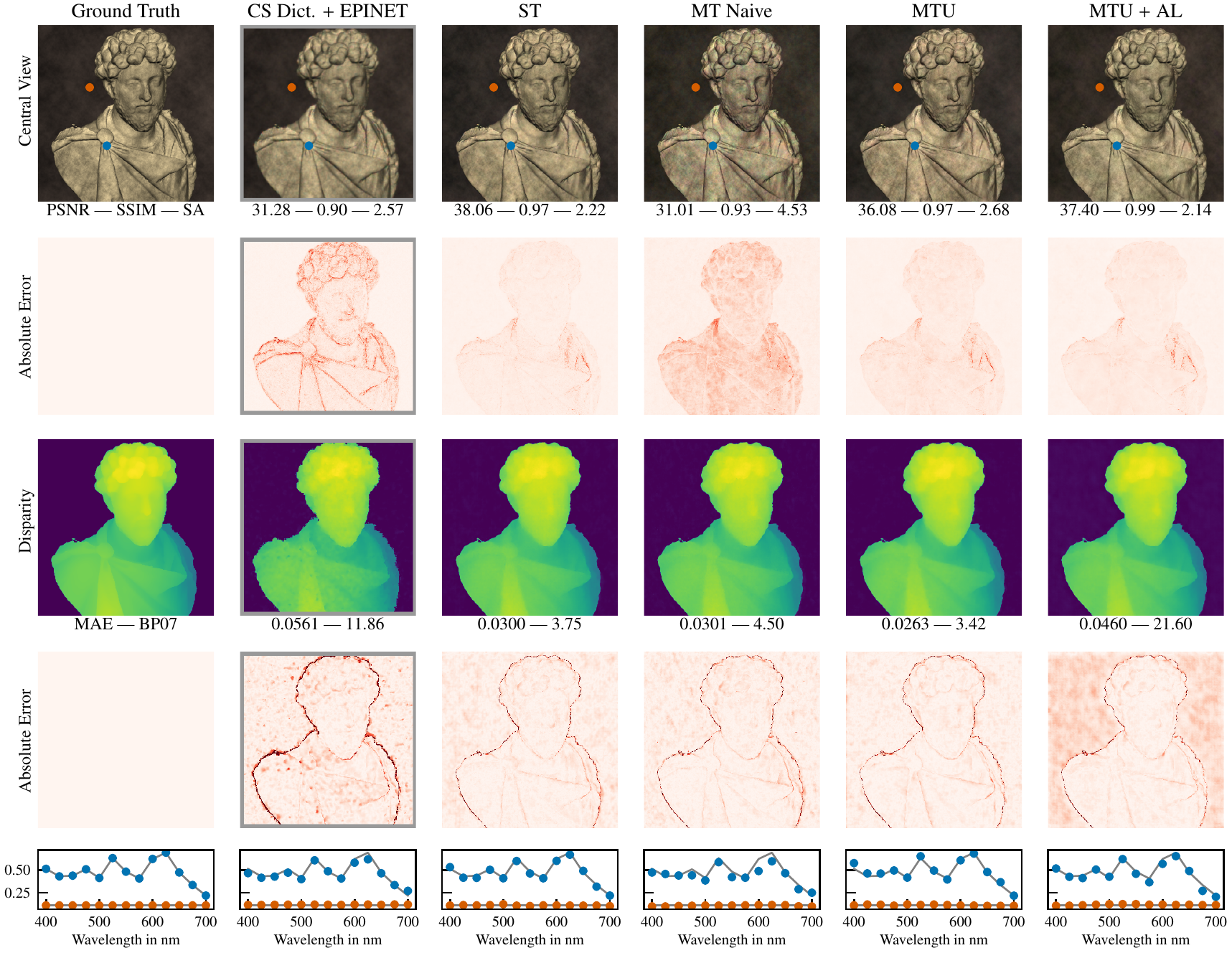}
    \caption{Performance comparison for full-sized prediction of the synthetic \textit{Bust} dataset challenge.
    The shown methods are Compressed Sensing (CS), Single Task (ST), as well as Multi Task (MT) methods.
    In the MT case we consider MT with uncertainty (MTU)~\cite{Kendall2018:MultiTaskUncertainty} and the proposed MTU approach using an auxiliary loss (MTU + AL).
    Evaluation metrics PSNR in dB, SA in $^\circ$, MAE in px and BP07 in \%.
    The spectra of the two marked reference points are depicted in blue and orange together with the ground truth spectrum in grey.}
    \label{fig:eval-challenge-bust}
\end{figure*}

\begin{figure*}
    \centering
    \includegraphics{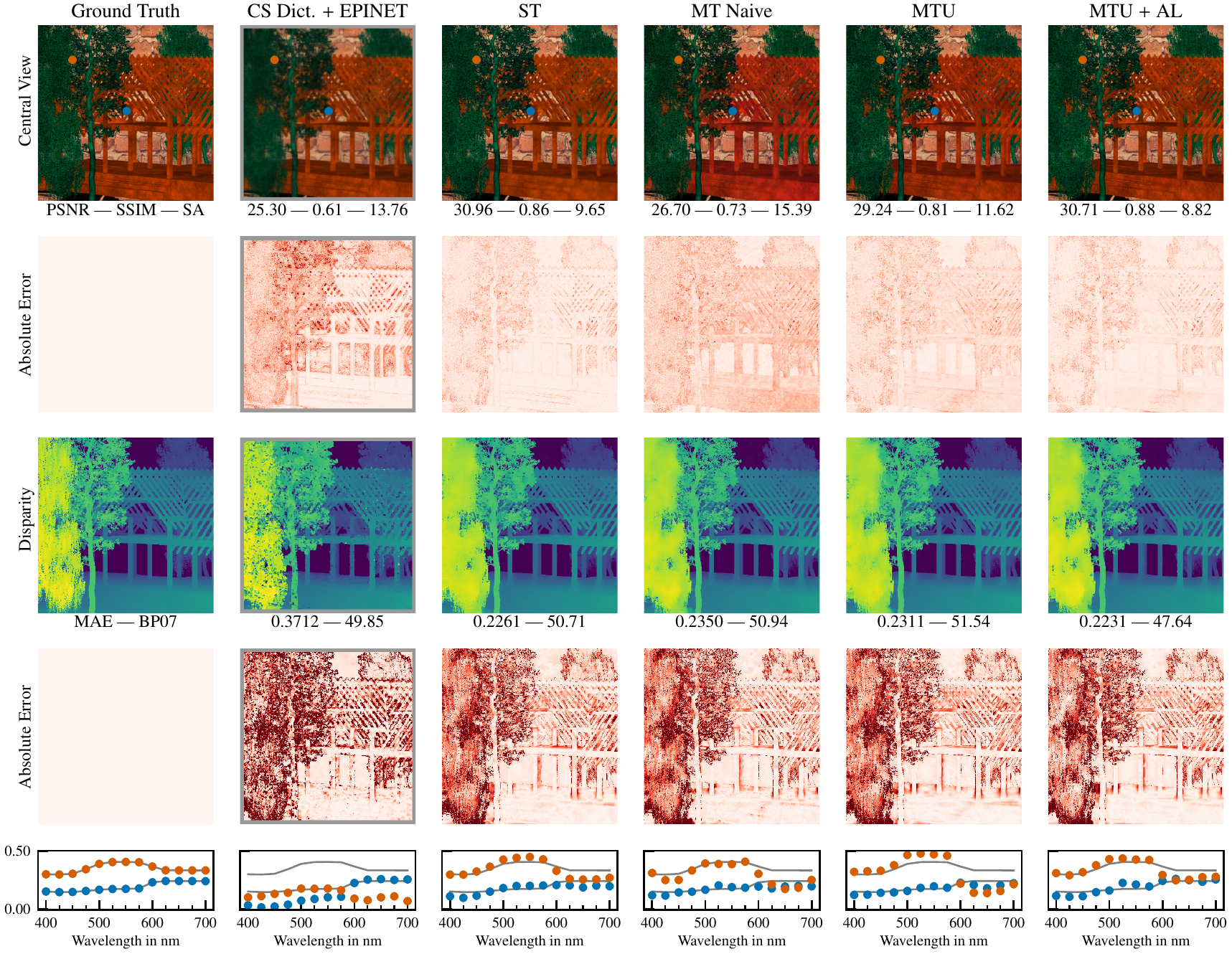}
    \caption{Performance comparison for full-sized prediction of the synthetic \textit{Cabin} dataset challenge.
    Captions are identical to Figure~\ref{fig:eval-challenge-bust}.}
    \label{fig:eval-challenge-cabin}
\end{figure*}

\begin{figure*}
    \centering
    \includegraphics{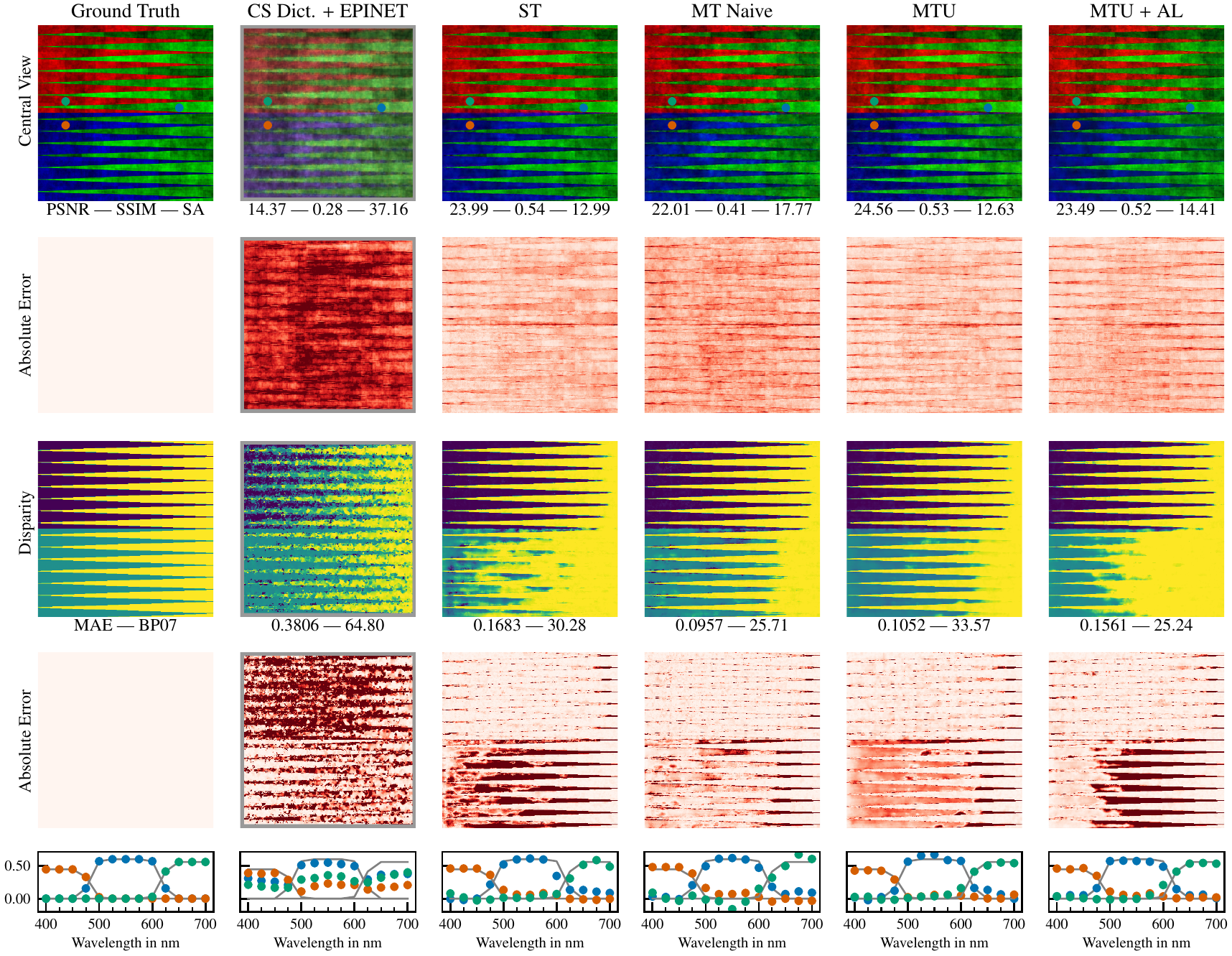}
    \caption{Performance comparison for full-sized prediction of the synthetic \textit{Backgammon} dataset challenge.
    Captions are identical to Figure~\ref{fig:eval-challenge-bust}.}
    \label{fig:eval-challenge-backgammon}
\end{figure*}

\begin{figure*}
    \centering
    \includegraphics{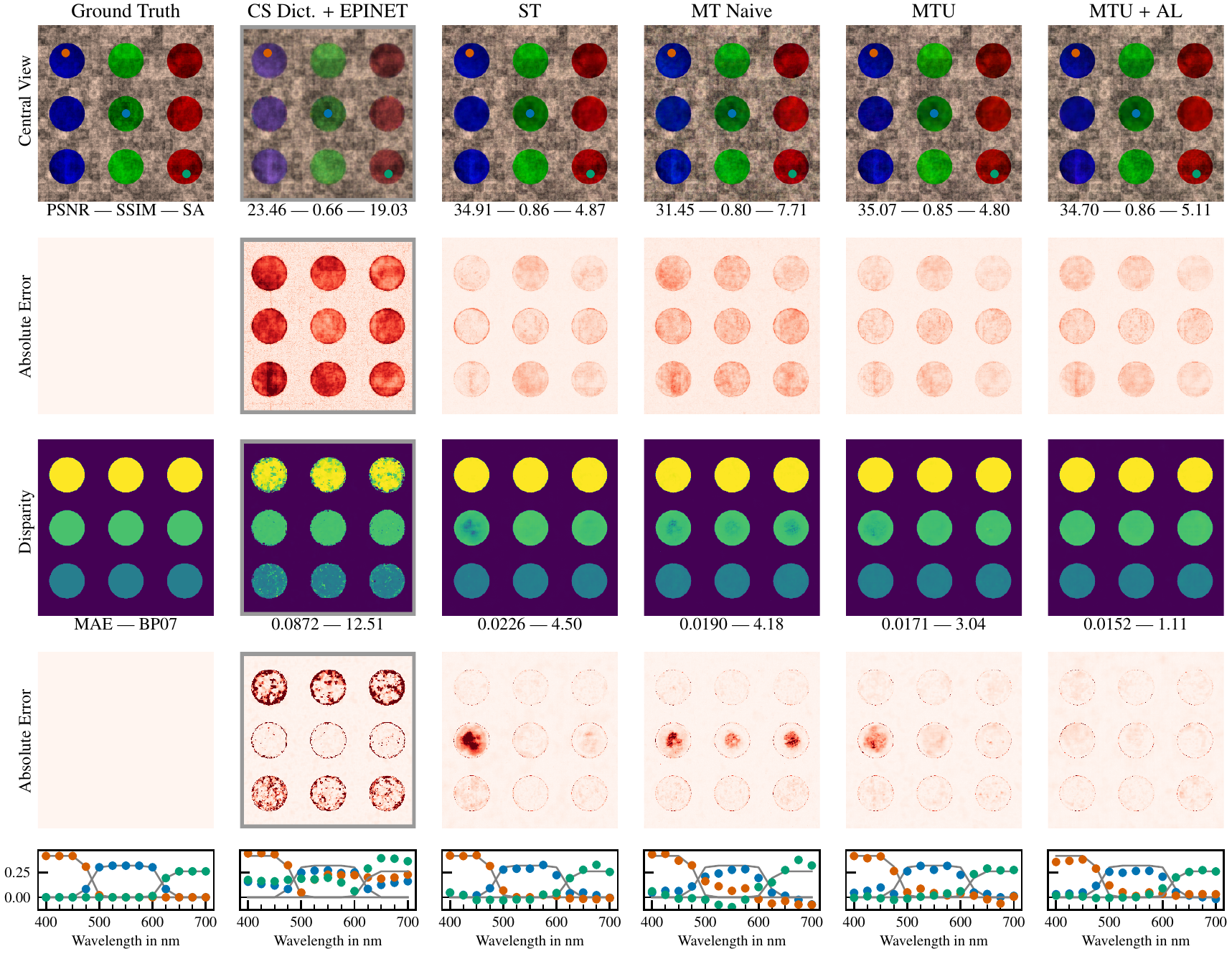}
    \caption{Performance comparison for full-sized prediction of the synthetic \textit{Circles} dataset challenge.
    Captions are identical to Figure~\ref{fig:eval-challenge-bust}.}
    \label{fig:eval-challenge-circles}
\end{figure*}

\begin{figure*}
    \centering
    \includegraphics{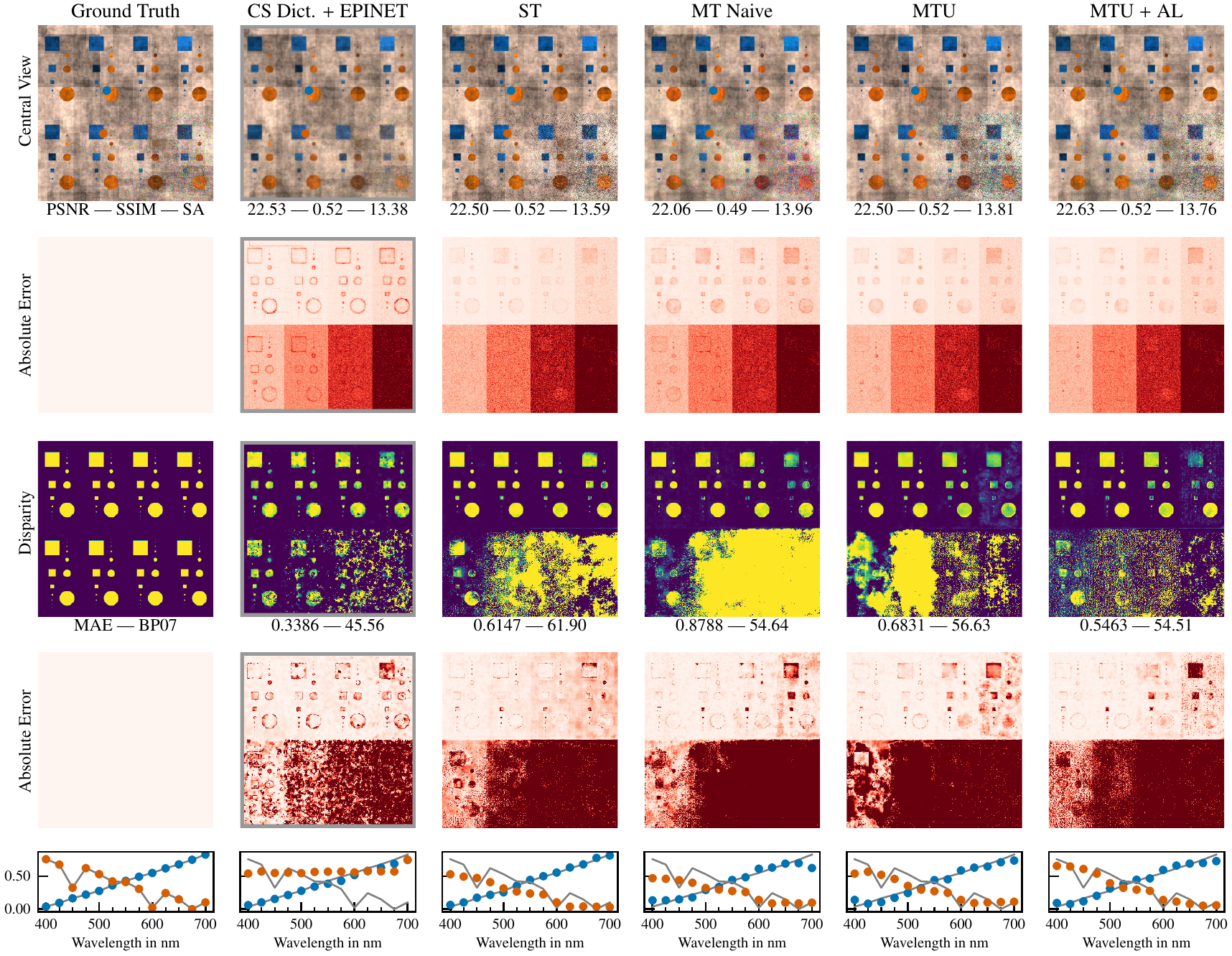}
    \caption{Performance comparison for full-sized prediction of the synthetic \textit{Dots} dataset challenge.
    Captions are identical to Figure~\ref{fig:eval-challenge-bust}.
    The scene is superposed with uncorrelated white noise resulting in a block-wise PSNR of 45\,dB in the top-left corner and decreasing in steps of 5\,dB to 10\,dB in the lower right corner.
    Hence, the mean reconstruction PSNR is not representative of the block wise reconstruction quality.
    }
    \label{fig:eval-challenge-dots}
\end{figure*}

\begin{figure*}
    \centering
    \includegraphics{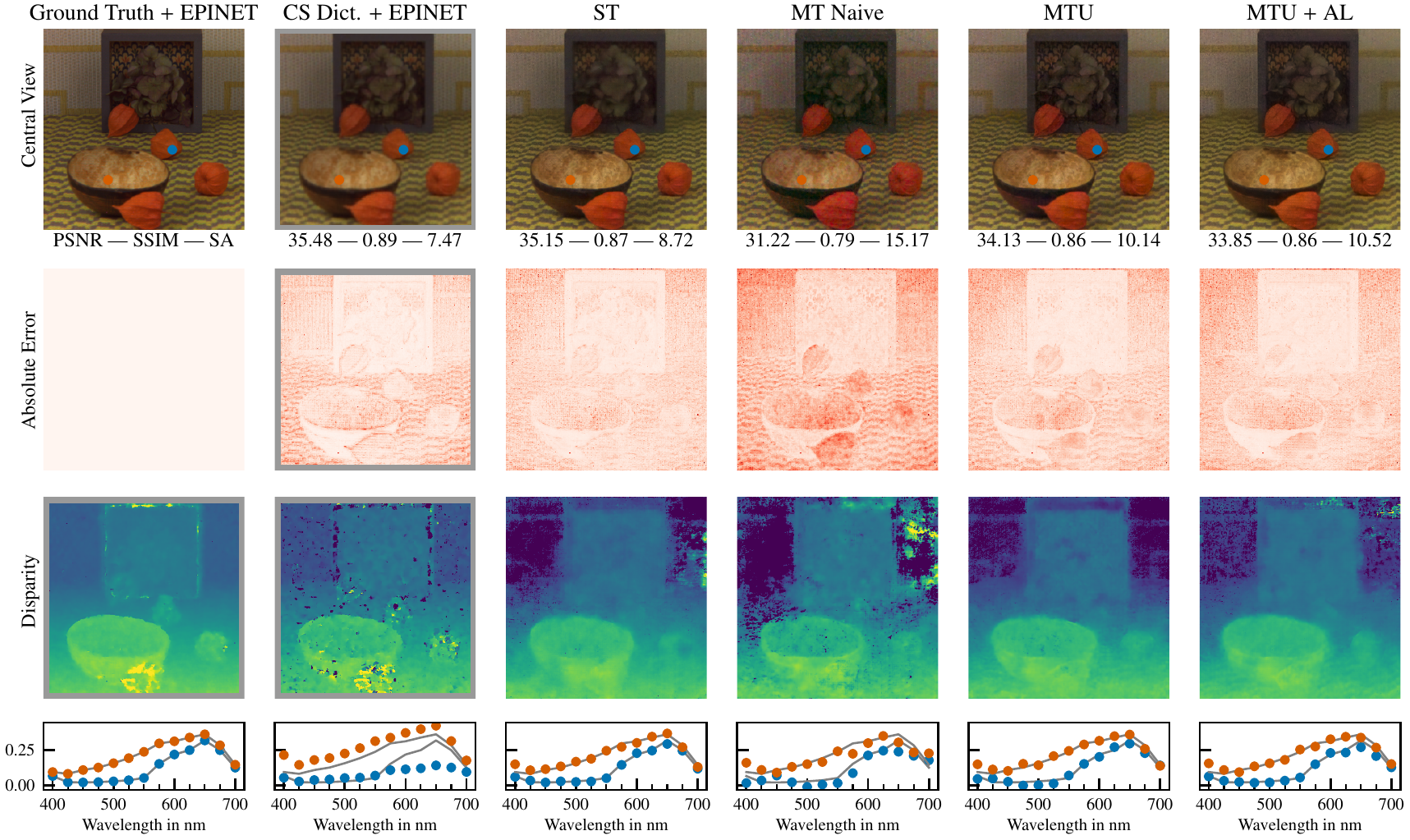}
    \caption{Performance comparison for full-sized prediction of the real-world \textit{Floral} scene.
    Captions are identical to Figure~\ref{fig:eval-challenge-bust}.}
    \label{fig:predict-real-floral}
\end{figure*}

\begin{figure*}
    \centering
    \includegraphics{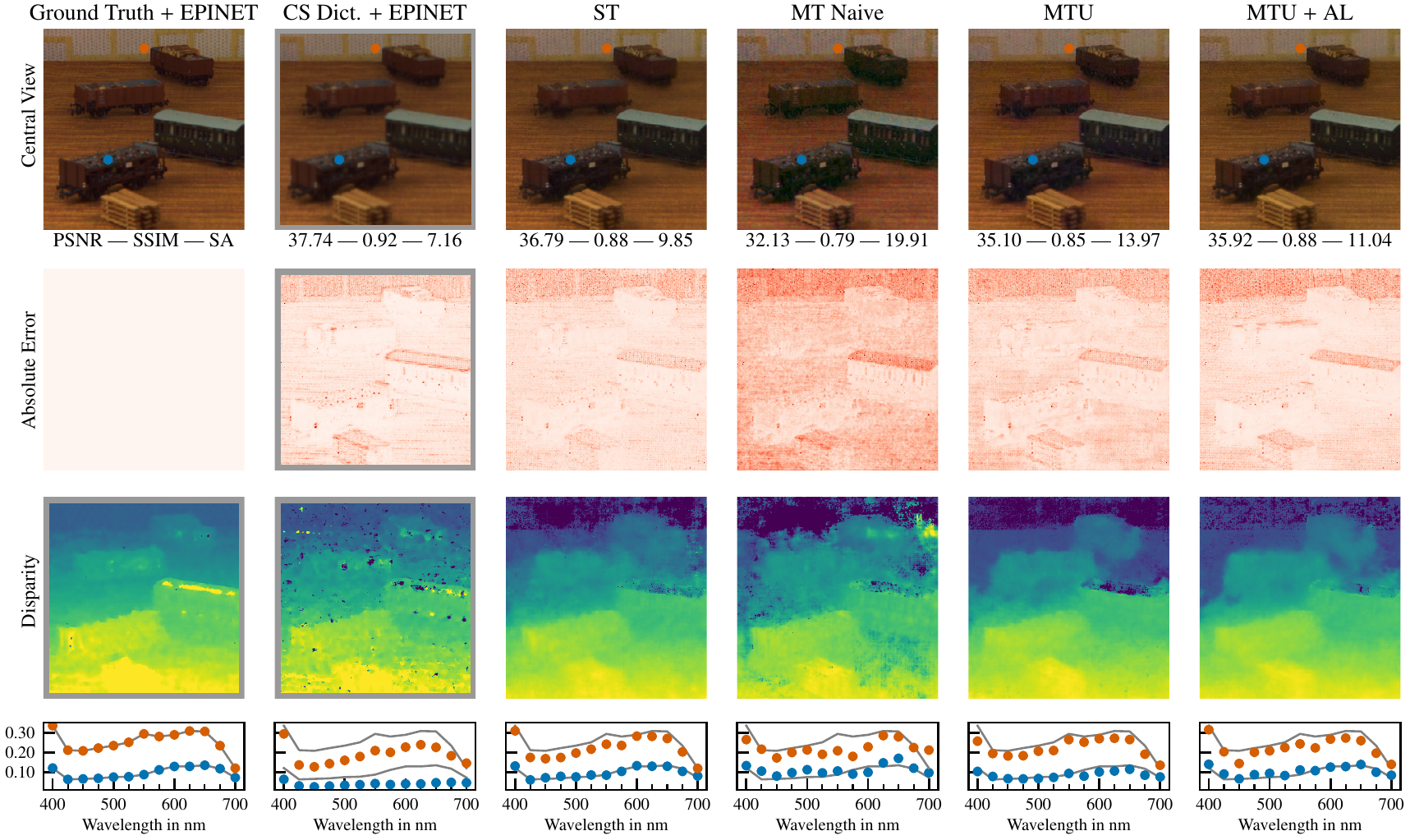}
    \caption{Performance comparison for full-sized prediction of the real-world \textit{Wagons} scene.
    Captions are identical to Figure~\ref{fig:eval-challenge-bust}.}
    \label{fig:predict-real-wagons}
\end{figure*}

\begin{figure*}
    \centering
    \includegraphics[]{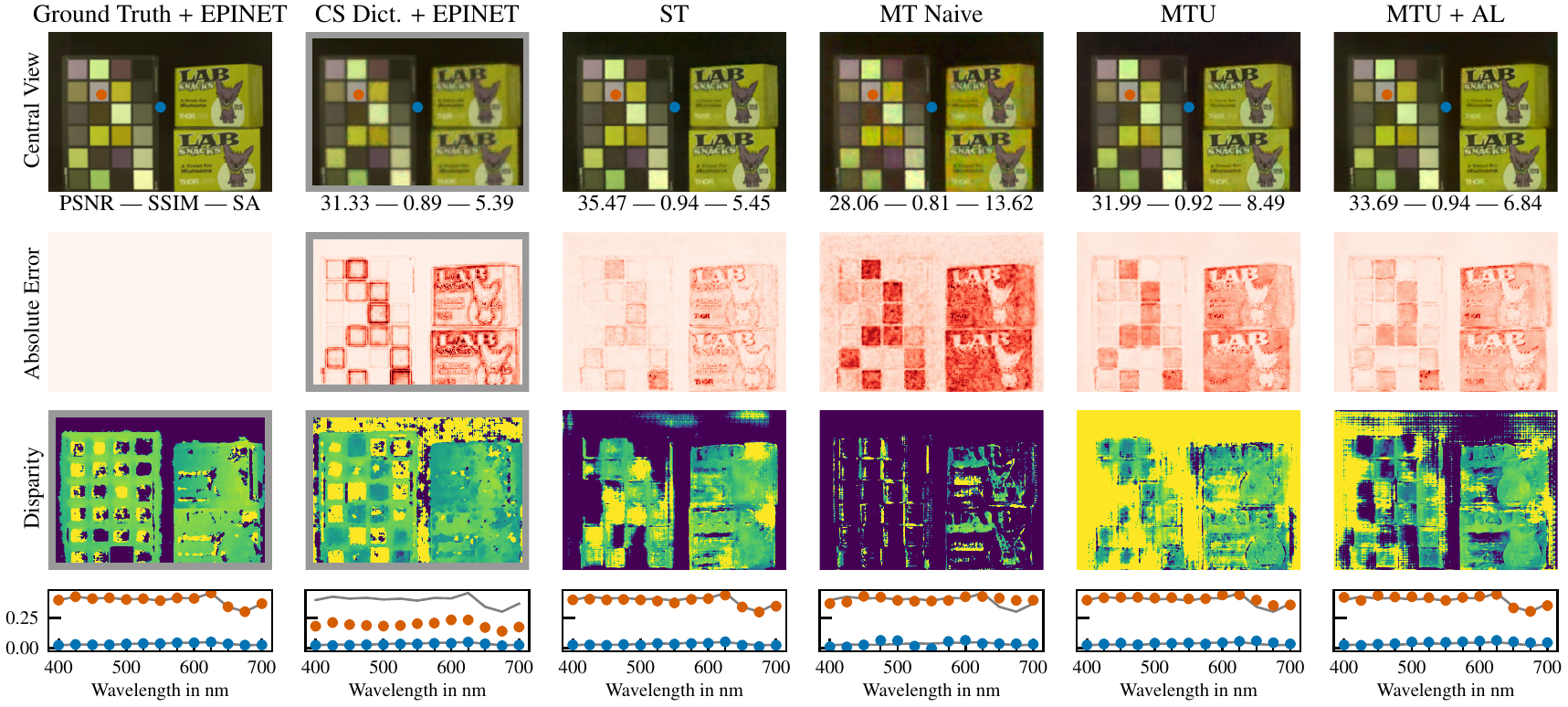}\\[2mm]
    \includegraphics[trim={0 0 0 3mm},clip]{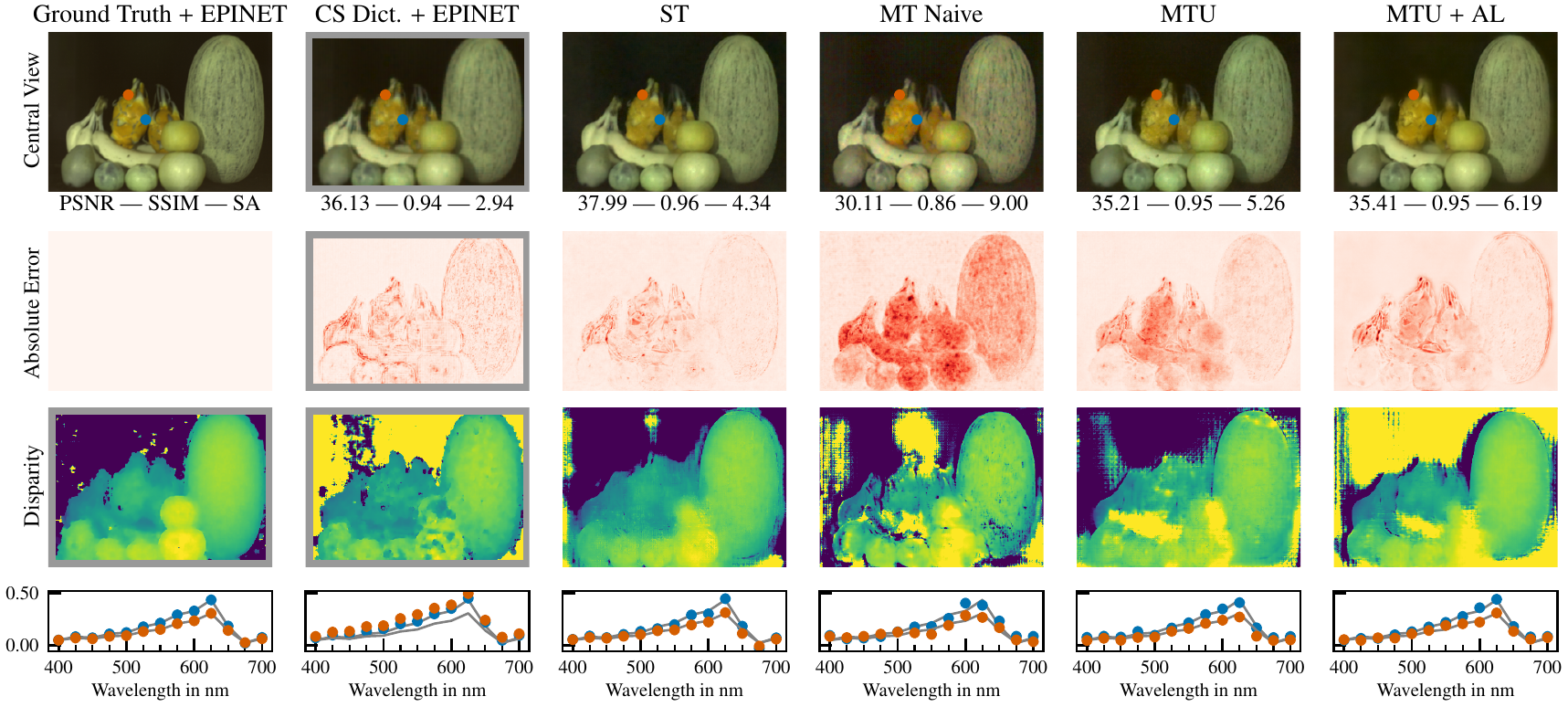}\\[2mm]
    \includegraphics[trim={0 0 0 3mm},clip]{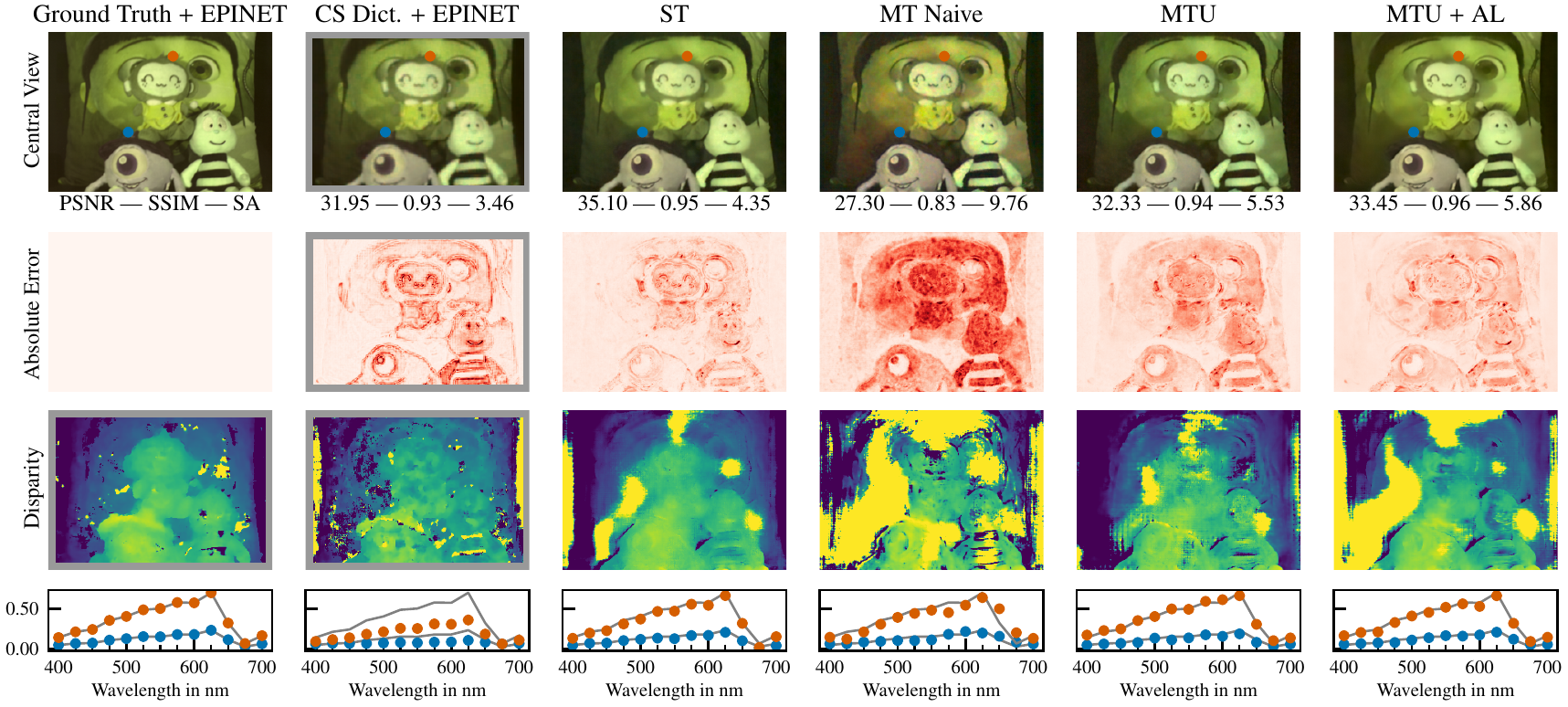}
    \caption{Performance comparison for full-sized prediction of the dateset by~Xiong~\etal~\cite{Xiong2017}.
    Captions are identical to Figure~\ref{fig:eval-challenge-bust}.}
    \label{fig:predict-real-xiong}
\end{figure*}

Additional evaluations of the synthetic data challenges are shown in Figures~\ref{fig:eval-challenge-bust}--\ref{fig:eval-challenge-dots}.
The remaining two scenes of the proposed real-world dataset are given in Figures~\ref{fig:predict-real-floral} and~\ref{fig:predict-real-wagons}.
Finally, the spectral light fields provided by Xiong~\etal~\cite{Xiong2017} are shown in Figure~\ref{fig:predict-real-xiong}.

%
%

\end{appendix}

\end{document}